\documentclass[runningheads]{llncs}
\usepackage{graphicx}
\usepackage{tikz}
\usepackage{comment}
\usepackage{amsmath,amssymb} 
\usepackage{color}
\usepackage{epsfig}
\usepackage{lipsum}
\usepackage{algorithm}
\usepackage{algpseudocode}
\usepackage{url}
\usepackage{xcolor, colortbl}
\usepackage{mathtools}
\usepackage{tabularx}
\usepackage{multirow}
\usepackage{multicol}
\usepackage{enumitem}
\usepackage{bbm}
\usepackage{wrapfig}
\usepackage{setspace}
\usepackage{rotating}
\RequirePackage{fix-cm}
\usepackage{booktabs}
\usepackage{kotex}
\usepackage{pifont}
\usepackage{blindtext}
\usepackage{xr}
\usepackage[toc]{multitoc}
\usepackage[pagebackref,breaklinks,colorlinks]{hyperref}
\newcolumntype{P}[1]{>{\centering\arraybackslash}p{#1}}
\usepackage[accsupp]{axessibility}  
\usepackage[capitalize]{cleveref}

\usepackage{soul}
\usepackage{makecell}
\usepackage{tabularx}
\usepackage{xcolor, colortbl}
\usepackage{enumitem}
\usepackage{subcaption}
\usepackage{adjustbox}
\usepackage{diagbox}
\usepackage{tabularx}
\usepackage{multirow}
\usepackage{pifont}
\usepackage{pict2e,picture}
\def\ie{\emph{i.e.}}
\def\eg{\emph{e.g.}}
\def\etal{\emph{et al.}}

\definecolor{grey}{rgb}{0.9, 0.9, 0.9}
\newcommand{\ccol}{\cellcolor{grey}}
\definecolor{brown}{rgb}{0.65, 0.16, 0.16}

\usepackage{pict2e,picture}

\makeatletter
\DeclareRobustCommand{\shortarrow}[1][]{%
  \check@mathfonts
  \if\relax\detokenize{#1}\relax
    \settowidth{\dimen@}{$\m@th\rightarrow$}%
  \else
    \setlength{\dimen@}{#1}%
  \fi
  \sbox\z@{\usefont{U}{lasy}{m}{n}\symbol{41}}%
  \begin{picture}(\dimen@,\ht\z@)
  \roundcap
  \put(\dimexpr\dimen@-.7\wd\z@,0){\usebox\z@}
  \put(0,\fontdimen22\textfont2){\line(1,0){\dimen@}}
  \end{picture}%
}
\makeatother

\usepackage[accsupp]{axessibility}  

\begin{document}
\pagestyle{headings}
\mainmatter
\def\ECCVSubNumber{4593}  

\title{Cross-Domain Ensemble Distillation \linebreak for Domain Generalization} 

\titlerunning{Cross-Domain Ensemble Distillation for Domain Generalization}
%
\author{Kyungmoon Lee%
\inst{1,2}
Sungyeon Kim\inst{1}
Suha Kwak\inst{1}
}
\authorrunning{Kyungmoon Lee, Sungyeon Kim, Suha Kwak}
%
\institute{$^{1}$POSTECH, Pohang, Korea \qquad $^{2}$NALBI Inc., Seoul, Korea \\
\email{kyungmoon@nalbi.ai, \{sungyeon.kim, suha.kwak\}@postech.ac.kr }\\
{\tt\small
\url{http://cvlab.postech.ac.kr/research/XDED/}
}
}

\maketitle

\begin{abstract}
Domain generalization is the task of learning models that generalize to unseen target domains. We propose a simple yet effective method for domain generalization, named cross-domain ensemble distillation (XDED), that learns domain-invariant features while encouraging the model to converge to flat minima, which recently turned out to be a sufficient condition for domain generalization. To this end, our method generates an ensemble of the output logits from training data with the same label but from different domains and then penalizes each output for the mismatch with the ensemble. Also, we present a de-stylization technique that standardizes features to encourage the model to produce style-consistent predictions even in an arbitrary target domain. Our method greatly improves generalization capability in public benchmarks for cross-domain image classification, cross-dataset person re-ID, and cross-dataset semantic segmentation. Moreover, we show that models learned by our method are robust against adversarial attacks and image corruptions.


\end{abstract}
\section{Introduction}

\begin{figure*}[t]
    \centering
    \includegraphics[width=0.9\textwidth]{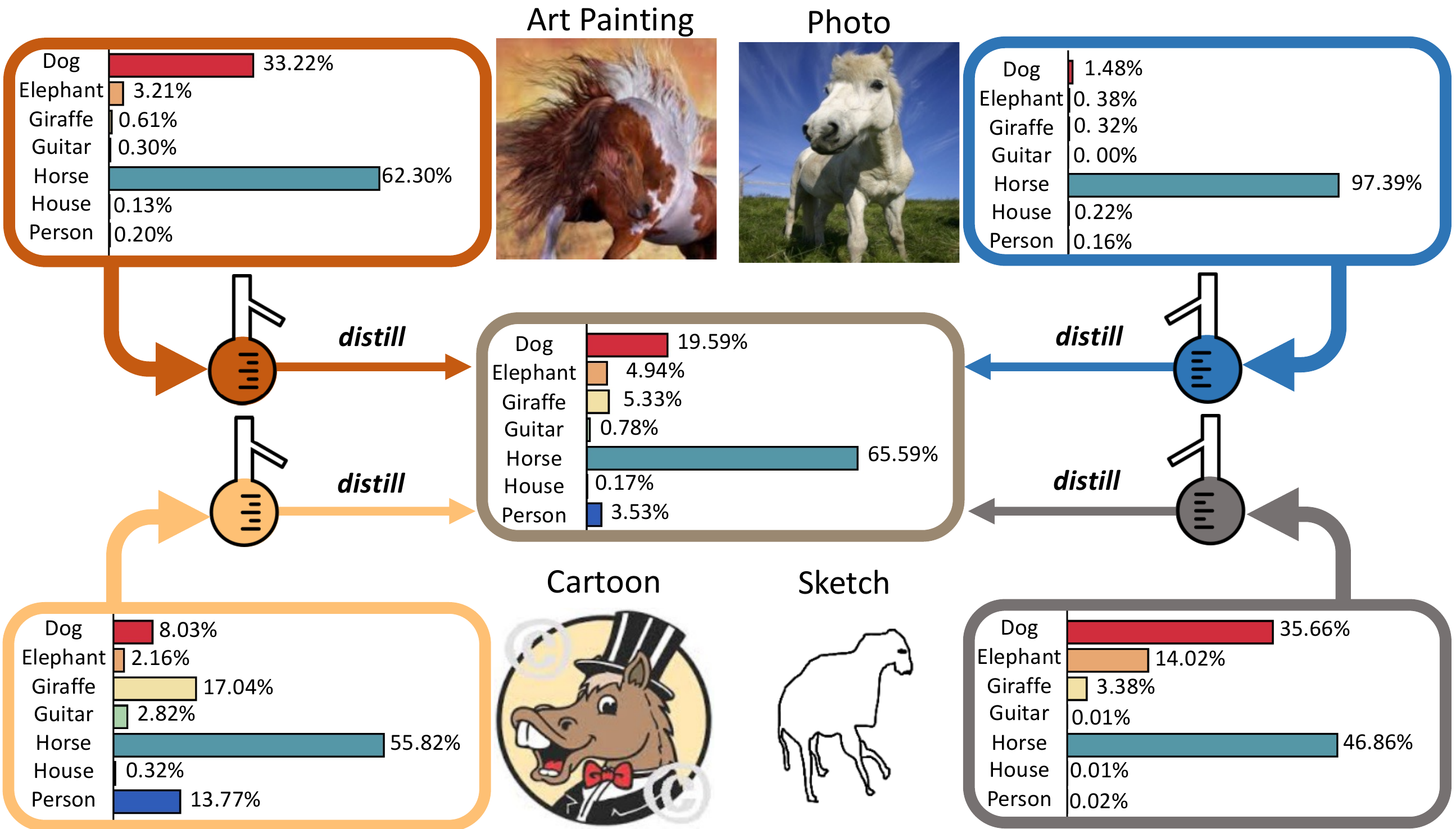}
    \caption{Illustration of cross-domain ensemble distillation (XDED). Although the four images share the same class label, their predictions manifest different inter-class relations due to the visual gap between domains. XDED constructs an ensemble by averaging all predictions and matches it with each prediction.
    }
    \label{fig:teaser}
\end{figure*}

Deep neural networks (DNNs) have brought remarkable advances in a number of research areas such as image classification~\cite{Alexnet}, image synthesis~\cite{goodfellow2014generative}, and reinforcement learning~\cite{mnih2013playing}. The huge success of DNNs depends heavily on the assumption that training and test data are sampled under the independent and identically distributed (i.i.d.)~condition. However, this assumption often does not hold in real-world scenarios; a large error occurs due to the discrepancy between training and test data, also known as the domain shift problem. As a solution to this problem, domain generalization, the task of learning models that generalize to unseen target domains, is in the spotlight.
A key to the success of domain generalization is to learn invariant features across domains. To this end, most previous methods align feature distributions of multiple domains by adversarial training~\cite{li2018domain,li2018deep}, minimizing the dissimilarity between the distributions of source domains~\cite{muandet2013domain}, or contrastive learning~\cite{kim2021selfreg}. Then, a classifier is trained to predict the labels for the aligned source features in hopes that it will also generalize well for any target domain. 
However, this approach often drops performance when the target domain differs substantially from the source domains as the model is prone to overfit to the source domains.

Meanwhile, the relationship between the geometry of loss landscapes and generalization ability has attracted increasing attention~\cite{dziugaite2017computing,foret2021sharpness,jiang2020fantastic,keskar2016large}.
In particular, converging to flat minima in loss landscapes is known as a key to achieve robustness against the
loss landscape shift between training and test datasets. 
Inspired by the observation that higher posterior entropy helps a model converge to flat minima~\cite{chaudhari2017entropy,pereyra2017regularizing,zhang2018deep}, entropy regularization techniques like self-knowledge distillation~\cite{zhang2019your} and entropy maximization~\cite{cha2020cpr} have been proposed to increase entropy rather than forcing a model to completely fit training data (\ie, one-hot labels) to induce low entropy.
Since the degree
of loss landscape shift is generally expected to be bigger in the case of domain generalization, it is more important to converge to flat minima 
in domain generalization.
However, the benefit of flat minima in terms of domain generalization has not been actively studied yet.

In this paper, we propose a novel method, named \emph{cross-domain ensemble distillation} (XDED), that learns domain-invariant features while encouraging convergence to flat minima for domain generalization.
Specifically, XDED generates an ensemble of the output logits for the data with the same label but from different domains, and then penalizes each output for the mismatch with the ensemble (Fig.~\ref{fig:teaser}). 
By doing so, it enables a model to learn domain-invariant features by enforcing prediction consistency between the data with the same label but from different domains. 
Also, 
XDED increases the posterior entropy of each output distribution, which helps the model converge to flat minima as the entropy regularization does.
To the best of our knowledge, XDED is the first to achieve these two objectives simultaneously for domain generalization, and this contribution leads to significant performance improvement.

Since XDED is still limited to exploiting the information of only source domains, there is further room to reduce the domain gap with the target domain. Hence, we also introduce a de-stylization technique well-suited to domain generalization, called UniStyle. UniStyle suppresses domain-specific style bias simply by standardizing intermediate feature maps of input image during both training and test time. Thanks to UniStyle, our model produces style-consistent predictions not only for the source domains but also for the target domain, which greatly reduces the domain gap and boosts the effect of XDED.

Based on the recent theoretical result on the relationship between the domain generalization and the flatness of local minima~\cite{cha2021swad}, we first empirically show that the proposed framework can improve generalization capability by achieving two goals: promoting flat minima and reducing the domain gap. Next, we further demonstrate the superiority of our method through extensive experimental results. 
On the standard public benchmarks for cross-domain image classification, XDED significantly enhances generalization ability in both multi-source and single-source settings. 
We also validate the effectiveness of our method in various domain generalization scenarios by showing the non-trivial improvement on the DomainBed~\cite{gulrajani2020search}, cross-dataset person re-ID~\cite{zheng2015scalable,zheng2017unlabeled}, and cross-dataset semantic segmentation experiments. Moreover, we demonstrate that models learned by our method also help achieve robustness against adversarial attacks and unseen image corruptions.

\section{Related Work}
\label{relatedwork}

\noindent \textbf{Domain generalization.} 
The goal of domain generalization is to learn domain-invariant features that well generalize to unseen target domains. 
For the purpose, existing methods match feature distributions of different domains by adversarial feature alignment~\cite{li2018domain,li2018deep} or reducing the difference between feature distributions of diverse source domains~\cite{muandet2013domain}. 
Recently, meta-learning frameworks~\cite{balaji2018metareg,dou2019domain,li2019episodic} have been introduced to simulate the domain shift by dividing the meta-train and meta-test domains from source domains. 
Also, data augmentation methods have been proposed to generate more diverse data beyond those of given source domains~\cite{StyleNeophile,kim2021wedge,shankar2018generalizing,xu2021fourier,zhou2020learning}. 
%
%
Most similar to our framework, ensemble methods for domain generalization have been proposed~\cite{xu2014exploiting,seo2020learning,zhou2021dael}. They all train multiple modules such as exemplar SVMs~\cite{xu2014exploiting}, domain-specific BN~\cite{Batchnorm} layers~\cite{seo2020learning} or classifiers~\cite{zhou2021dael}, and exploit the ensemble of learned modules for prediction in testing.
However, we remark that our XDED utilizes the ensemble of model predictions as the soft label and transfers it to the model itself. Therefore, it does not demand any additional module during both training and testing. 

\noindent \textbf{Knowledge distillation (KD).}
KD
was originally studied to transfer the knowledge of a deep model to a shallow model for model compression~\cite{hinton2015distilling}. 
%
It has been also used for other purposes such as metric learning~\cite{park2019relational,kim2021embedding} and network regularization~\cite{xu2019data,zhang2019your,yun2020regularizing}.
In particular for network regularization, self-knowledge distillation (self-KD) has been studied; it distills knowledge from the model itself and enforces prediction consistency between a sample and its perturbed one or other samples.
KD has been used for domain adaptation~\cite{meng2018adversarial,feng2020kd3a}, and such method trains several teacher models from the source domains and distills the ensemble of their predictions to the student model. 
It unfortunately requires large memory due to multiple teachers, and are difficult to be extended to domain generalization as they demand target images in training.
In contrast, our method improves generalization capability of a model on unseen domains without the need for target images and additional teacher models.

\noindent \textbf{Flat minima in loss landscapes.}
Recent analyses have revealed that finding flat minima is crucial for model generalization~\cite{keskar2016large,dziugaite2017computing,foret2021sharpness}. 
In this context,
multiple methods have been proposed to promote flat minima in loss landscapes since flat minima have an advantage over sharp minima in robustness against the loss landscape shift between training and test data.
Among literature on ways of promoting flat minima (\eg, weight averaging~\cite{izmailov2018averaging,cha2021swad} and training strategies~\cite{foret2021sharpness,chaudhari2017entropy}), we focus on the high entropy-seeking approaches, on which XDED is based. 
Maximum Entropy~\cite{pereyra2017regularizing,cha2020cpr} maximizes the entropy of an output distribution from a classifier.
Similarly, KD-based methods also aim at inducing high entropy of the output distribution by penalizing the mismatch with the output distribution from that of another classifier such as differently initialized peer networks~\cite{zhang2018deep} or subnetworks within a network itself~\cite{zhang2019your}. Although SWAD~\cite{cha2021swad} has introduced the importance of flat minima in the area of domain generalization, we remark that SWAD belongs to weight averaging but does not focus on learning domain-invariant features, whereas XDED belongs to entropy regularization as well as is designed for learning domain-invariant features.

\noindent \textbf{Bias towards styles.}
Recent studies~\cite{brendel2019approximating,geirhos2018imagenet} revealed that DNNs overly depend on a strong bias towards styles, and it is also confirmed in the domain generalization literature~\cite{choi2021robustnet,zhou2021domain,StyleNeophile} that a visual domain is highly correlated to feature statistics.
Hence, previous work defines image styles as the bias and attempts to remove the bias by style augmentation in the space of feature statistics~\cite{zhou2021domain,StyleNeophile}, using another model that is intentionally biased to styles~\cite{nam2021reducing}, or minimizing a whitening loss~\cite{choi2021robustnet}.
Distinct from these techniques, we show that a simple yet effective de-stylization technique leads to a smaller divergence measure between target and source domains without bells and whistles.


\section{Our Method}
\label{method}
\subsection{Cross-Domain Ensemble Distillation}
\noindent \textbf{Review of knowledge distillation (KD).}
The goal of KD~\cite{hinton2015distilling} is to transfer knowledge of a teacher model $t$ to a student model $s$, usually a wide and deep model to a smaller one, for the purposes of model compression or model regularization. Given input data point $x$ and its label $y \in \{1, \cdots, C \} $, we denote the output logit of model as $z(x) = [z_{1}(x), \cdots, z_{C}(x)]$. The posterior predictive distribution of $x$ is then formulated as:
\begin{align}
    \begin{split}
    P(y |x; \theta, \tau) = \frac{\textrm{exp}( z_{y}(x) / \tau ) }
    {\sum^{C}_{i=1}\textrm{exp}(z_{i}(x)/\tau)},
    \end{split}
    \label{eq:softmax}
\end{align}
where the model is parameterized by $\theta$ and $\tau$ is a temperature scaling parameter. KD enforces to match the predictive distributions of $s$ and $t$. Specifically, it is achieved by minimizing the Kullback-Leibler (KL) divergence between their predictive distributions as follows:
\begin{align}
    \begin{split}
    \mathcal{L}_{\textrm{KD}}(X; \theta_{s},\tau) = \sum_{x_{i} \in X}
    \sum^{C}_{c=1}
    D_{KL}(
    P(c|x_{i}; \theta_{t}, \tau) ||
    P(c|x_{i}; \theta_{s}, \tau)
    ),
    \end{split}
    \label{eq:original_kd}
\end{align}
where $X$ is a batch of input data, $\theta_{t}$ and $\theta_{s}$ are the parameters of a teacher and a student, respectively.

\noindent \textbf{Cross-domain ensemble distillation.} We propose a new KD method for domain generalization, called cross-domain ensemble distillation (XDED). XDED aims to construct the domain-invariant knowledge from the data of multiple domains. Specifically, XDED generates an ensemble of logits from the data with the same label but from different domains. Next, XDED penalizes each logit for the mismatch with the ensemble which is not biased towards a specific domain, which encourages learning domain-invariant features.
Unlike the conventional KD, XDED does not require an additional network that increases training complexity (\eg, extra parameters and training time) but distills the ensemble constructed by multiple samples to the model itself in the form of self-KD. 

Formally, let $X_{y}$ denote the set of samples that have the same class label $y$ in a mini-batch. Then, we obtain an ensemble of logits from $X_{y}$ by simply taking an average as:
\begin{align}
    \begin{split}
    \bar{z}(X_{y}) = \sum_{x_{i} \in X_{y}} \frac{ z(x_{i}) }{|X_{y}|}.
    \end{split}
    \label{eq:ensemble_logit}
\end{align}
Then, the predictive distribution for the ensemble created from data $X_{y}$ is as:
\begin{align}
    \begin{split}
    \bar{P}(c | X_{y} ; \theta, \tau) = \frac{\textrm{exp}( \bar{z}_{c}(X_{y}) / \tau ) }
    {\sum^{C}_{i=1}\textrm{exp}( \bar{z}_{i}(X_{y})/\tau)},
    \end{split}
    \label{eq:softmax}
\end{align}
The loss function of XDED is defined as follows:
\begin{align}
    \begin{split}
    \mathcal{L}_{\textrm{XDED}}(X_{y}; \theta, \tau) = \sum_{x_{i} \in X_{y}}
    \sum^{C}_{c=1}
    D_{KL}(
    \bar{P}(c|X_{y}; \hat{\theta}, \tau) ||
    P(c|x_{i}; \theta, \tau)
    ),
    \end{split}
    \label{eq:xe_ed}
\end{align}
where $\hat{\theta}$ is a fixed copy of the parameter $\theta$. Following \cite{miyato2018virtual,yun2020regularizing}, we stop the gradient to be propagated through $\hat{\theta}$ to prevent the model from falling into some trivial solutions. To sum up, we set our objective function as
\begin{align}
    \begin{split}
    \min_{\theta}L_{\theta} = 
    \mathcal{L}_{\textrm{CE}}(X,Y; \theta) + \lambda \sum_{y \in \{Y\} } \mathcal{L}_{\textrm{XDED}}(X_{y}; \theta, \tau),
    \end{split}
    \label{eq:total_loss}
\end{align}
where $X$ is a batch of input images, $Y$ is a batch of corresponding class labels, $\mathcal{L}_{\textrm{CE}}$ denotes the vanilla cross-entropy loss, and $\lambda$ is a hyperparameter to balance $\mathcal{L}_{\textrm{CE}}$ and $\mathcal{L}_{\textrm{XDED}}$. $\lambda$ and $\tau$ are 5.0 and 4.0 throughout this paper.

\subsection{UniStyle: removing and unifying style bias} To further regularize the model to produce style-consistent predictions, we propose a de-stylization technique that is well-suited to domain generalization. As source domain styles are not expected to appear at test time, we propose UniStyle to prevent the model from being biased towards the domain-specific styles, which reduces the domain gap with the target domain.

More specifically, following existing methods based on style transfer~\cite{dumoulin2016learned,huang2017arbitrary,ulyanov2016instance}, we first represent a neural style as statistics of intermediate feature maps from the feature extractor. Formally, let $ F \in \mathbb{R}^{C \times H \times W} $ denote an  intermediate feature map of an image. Then, a neural style of the image is represented as the combination of channel-wise mean $\mu(F) \in \mathbb{R}^{C}$ and standard deviation $ \sigma(F) \in \mathbb{R}^{C} $ of $F$ as:
\begin{equation}
\mu_{c}(F) = \frac{1}{HW} \sum^{H}_{h=1}\sum^{W}_{w=1} F_{c,h,w},
\end{equation}
and
\begin{equation}
\sigma_{c}(F) = \sqrt{ \frac{1}{HW} \sum^{H}_{h=1}\sum^{W}_{w=1} ( F_{c,h,w} - \mu_{c}(F) )^{2} },
\end{equation}
where $\mu(F) = [ \mu_{1}(F), \cdots, \mu_{C}(F) ] $ and $\sigma(F) = [ \sigma_{1}(F), \cdots, \sigma_{C}(F) ] $. Next, we simply standardize each feature to have constant channel-wise statistics, $ \mu_{W} $ and $\sigma_{W} $ as:
\begin{equation}
\textrm{UniStyle}(F) = \sigma_{W} \frac{F-\mu(F)}{\sigma(F)} + \mu_{W},
\end{equation}
where $\mu_{W} = \mathbf{0} $ and $ \sigma_{W}= \mathbf{1} $ (\ie, zero-mean standardization). 
Technically, UniStyle is a special case of InstanceNorm (IN)~\cite{ulyanov2016instance}. Nevertheless, we remark that UniStyle aims to remove domain-specific information without any learnable parameters to reduce the domain gap while IN learns channel-wise scaling and bias parameters for style transfer.
Also, note that we empirically observed that UniStyle is effective when being applied at multiple early layers, which is aligned with recent studies~\cite{dumoulin2016learned,huang2017arbitrary} suggesting that the style information is usually captured at the early layers.\footnote{See the supplementary material for further analyses.}


\subsection{Analysis of Our Method}
In this section, we analyze the effectiveness of XDED, especially through the link to the theoretical result and the supporting empirical evidences. We first begin with a theorem related to domain adaptation~\cite{ben2010theory,ben2007analysis}, which shows that the expected risk on the target domain is bounded by that on the source domain and the divergence between these domains.
To find a model parameter $\theta \in \Theta$ for domain generalization, Cha \etal~\cite{cha2021swad} considered a robust empirical loss:
\begin{align}
    \begin{split}
    \hat{\varepsilon}^{\gamma}_{S}(\theta) := \max_{|| \Delta || \leq \gamma} \hat{\varepsilon}_{S}(\theta+\Delta)
    \end{split}
    \label{eq:robust_risk}
\end{align}
where $\hat{\varepsilon}_{S}(\theta)$ is an empirical risk over source domains $S$ and $\gamma$ is a radius which defines neighbor parameters of $\theta$. Then, Cha \etal~\cite{cha2021swad} proved that finding flat minima reduces the domain gap through the theorem below:
\begin{theorem}
Consider a set of N covers $ \{ \Theta_{k} \}^{N}_{k=1} $ such that the hypothesis space $ \Theta \subset \cup^{N}_{k}\Theta_{k}$ where $ diam(\Theta) := \sup_{\theta,\theta' \in \Theta} || \theta - \theta' ||_{2}, N:= \lceil (diam(\Theta)/\gamma)^{d} \rceil $ and $d$ is dimension of $\Theta$. Let $v_{k}$ be a VC dimension of each $\Theta_{k}$. Then, for any $\theta \in \Theta$, the following bound holds with probability at least $1-\delta$,
    \begin{align}
        \varepsilon_{T}(\theta) <  & \hat{\varepsilon}^{\gamma}_{S}(\theta) + \frac{1}{2I} \sum^{I}_{i=1} \textrm{Div}(S_{i},T) + \max_{k\in[1,N]}  \sqrt{ \frac{v_{k}\ln{(m/v_{k})} + \ln{(N/\delta)}}{m}},
        \label{eq:swad}
    \end{align}
where $m=nI$ is the number of training samples and $ \textrm{Div}(S_{i},T) $ is the divergence between the source domain $S_{i}$ and the target domain $T$.
\label{theorem:swad}
\end{theorem}
We remark that, in Eq.~(\ref{eq:swad}), the test loss $\varepsilon_{T}(\theta)$ is bounded by three terms: (1) the robust empirical loss $\hat{\varepsilon}^{\gamma}_{S}(\theta)$, (2) the divergence $\textrm{Div}(S_{i}, T)$, and (3) a confidence bound depending on the radius $\gamma$ and the number of training samples $m$. In the rest of this section, according to the above theorem, we provide a theoretical interpretation that our method enhances the generalization ability by lowering both $\hat{\varepsilon}^{\gamma}_{S}(\theta)$ and $\textrm{Div}(S_{i}, T)$ with the empirical evidences. 

\begin{table}[!t]
    \centering
    \caption{
    Comparison of the entropy values. When each model is converged, the entropy value is calculated by averaging over all training samples.
    }
    \fontsize{8.5}{10.5}\selectfont
    \begin{tabularx}{0.7 \textwidth}{
       >{\centering\arraybackslash}X|
       >{\centering\arraybackslash}X
       >{\centering\arraybackslash}X|
       >{\centering\arraybackslash}X
       >{\centering\arraybackslash}X}
    
    \hline
    \multicolumn{1}{l|}{} &
    \multicolumn{2}{c|}{OfficeHome (Clipart)} & \multicolumn{2}{c}{PACS (Cartoon)} \\
    \multicolumn{1}{l|}{Methods} & {Entropy} & {Accuracy} & {Entropy} & {Accuracy} \\
    
    \hline
    
    \multicolumn{1}{l|}{ResNet-18} & 0.25 & 49.4 & 0.01 & 75.9\\
    \multicolumn{1}{l|}{MixStyle~\cite{zhou2021domain}} & 0.35 & 53.4 & 0.03 & 78.8 \\
 
    \multicolumn{1}{l|}{\ccol XDED} & \ccol \textbf{0.92} & \ccol \textbf{55.2} & \ccol \textbf{0.38} & \ccol \textbf{81.7} \\
        \hline
    \end{tabularx}
    \label{tab:supple_comparison_entropy}
\end{table}
\begin{figure*}[t]
    \centering
    \includegraphics[width=0.85\textwidth]{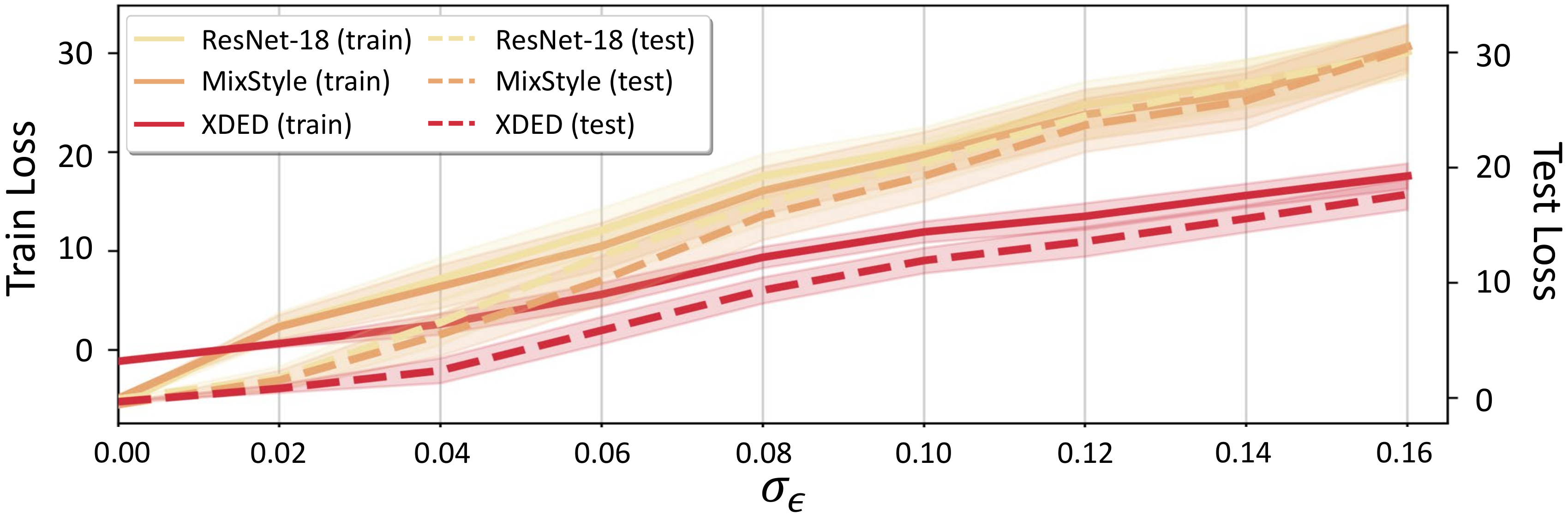}
    \caption{
    Train/Test losses versus the weight perturbation while varying the standard deviation of the added Gaussian noise. Note that the results are produced with the target domain (Art of PACS) and the rest source domains, and the loss values are log-scaled.
    }
    \label{fig:a_dist}
\end{figure*}

\begin{figure*}[t]
    \centering
    \includegraphics[width=0.9\textwidth]{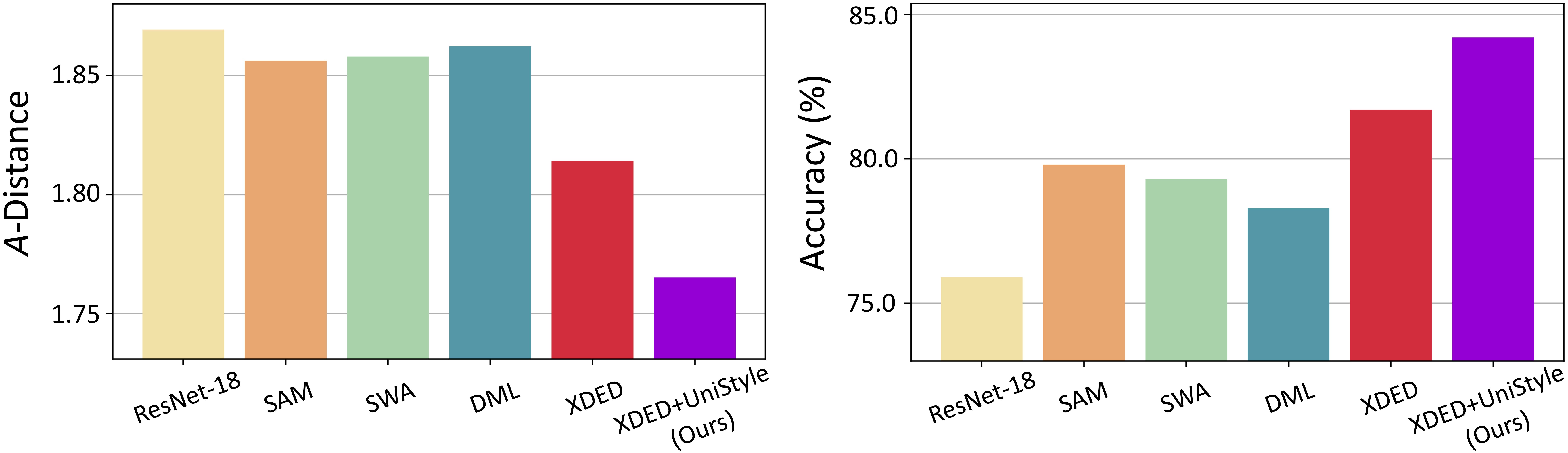}
    \caption{
    Comparison to existing methods promoting flat minima. Each model is evaluated on Cartoon of PACS after being trained on the rest source domains. \textbf{Left}: The divergence ($\mathcal{A}$-distance) between the source domains and the target domain, \textbf{Right}: Generalization performance on the target domain.
    }
    \label{fig:supple_a_dist}
\end{figure*}

\noindent \textbf{Promoting flat minima.} 
We remark that XDED is motivated by recent entropy regularization methods~\cite{cha2020cpr,zhang2019your,zhang2018deep} in pursuit of flat minima. It has been empirically demonstrated that these methods promote flat minima by inducing higher posterior entropy. It can be interpreted as relaxing the training procedure to learn richer information encoded in soft labels, which helps the model converge to flat minima more than forcing the model to completely fit one-hot labels. In this context, we also demonstrate that XDED clearly induces higher entropy as shown in Table~\ref{tab:supple_comparison_entropy}. Considering that XDED is motivated by the observation that different domains manifest different inter-class relations due to the domain gap (Fig.~\ref{fig:teaser}), this is natural since our ensembles would integrate meaningful inter-class relations from multiple domains and the model learned with them would be led towards high entropy.

Next, to investigate whether the model learned with XDED converges to flat minima indeed, we quantify the flatness of the local minima where the model converged by measuring the increase of loss values between $\theta$ and its neighborhoods, assuming that the model converged in flat minima would have smaller increases. Following \cite{cha2021swad,cha2020cpr,zhang2019your,zhang2018deep}, we measure the losses of the learned models before and after adding Gaussian noises to model parameters while varying the standard deviation of the noise $\sigma{}_{\epsilon}$ (\ie, $\mathcal{L}_{\textrm{CE}}(X,Y; \theta + \epsilon )$ where $ \epsilon \sim N(0, \sigma{}_{\epsilon}) $) with 100 runs. As a result, XDED demonstrates its robustness against the weight perturbation with smaller loss increases as shown in Fig.~\ref{fig:a_dist}. 

\noindent \textbf{Domain-invariant feature learning.} 
Here, we highlight that XDED also learns domain-invariant features via regularizing the consistency between the predictions from the data with the same label but from different domains and their ensemble. Thus, we compare XDED with existing methods promoting flat minima, which are dedicated to the flatness of local minima only. Specifically, to examine the effectiveness in reducing the divergence $\textrm{Div}(S_{i},T)$, we measure $\mathcal{A}$-distance~\cite{ben2010theory,kifer2004detecting}. Due to the computational intractability, we calculated an approximated one~\cite{long2015learning,nam2021reducing}~\footnote{It is defined as $ \hat{d}_{\mathcal{A}} = 2(1-2 \epsilon_{\textrm{svm}})$ where $\epsilon_{\textrm{svm}}$ is the generalization error of a SVM-based two-class classifier trained to distinguish between target and source domains.}
As shown in Fig.~\ref{fig:supple_a_dist} (Left), we observe that the existing methods promoting flat minima fail to reduce the distance while XDED clearly lowers the distance and UniStyle further enhances the result. Naturally, that result is connected to the quantitative superiority of our framework over existing flat minima-promoting methods (Fig.~\ref{fig:supple_a_dist} (Right)). 

\section{Experiments}
\label{experiments}
\subsection{Generalization in image classification}
\label{dg_cls}

\noindent \textbf{Multi-source domain generalization.}
Specifically, for a fair comparison, we follow the leave-one-domain-out protocol~\cite{li2017deeper} where we train a model on three domains and evaluate it on the remaining domain. For the benchmark datasets, we employ the PACS~\cite{li2017deeper} and OfficeHome~\cite{venkateswara2017deep} that are widely-used benchmarks for domain generalization in image classification. PACS contains 9,991 images of 7 classes over 4 domains: Art Painting, Cartoon, Photo, and Sketch. OfficeHome includes 15,500 images of 65 classes over 4 domains: Artistic, Clipart, Product, and Real. We use ResNet-18~\cite{resnet} as the backbone, and our UniStyle is applied to output feature maps of the first and second residual blocks for PACS and the first one only for OfficeHome.

\noindent \textbf{Results.}
As summarized in Table.~\ref{tab:cr_pacs_officehome}, we observe that our method not only significantly enhances the vanilla but also outperforms the latest competing methods. In particular, our method outperforms the second-best method on Cartoon of PACS and Clipart of OfficeHome by about 4.0\% and 2.0\%, respectively. these results justify the superiority of our method, which is simple yet effective.

\begin{table}[!t]
    \centering
    \caption{
    Leave-one-domain-out generalization results on PACS and OfficeHome.
    }
    \fontsize{8}{10}\selectfont
    \begin{tabularx}{1.0 \textwidth}{
       >{\centering\arraybackslash}X|
       >{\centering\arraybackslash}X
       >{\centering\arraybackslash}X
       >{\centering\arraybackslash}X
       >{\centering\arraybackslash}X
       >{\centering\arraybackslash}X
       >{\centering\arraybackslash}X
       >{\centering\arraybackslash}X
       >{\centering\arraybackslash}X
       >{\centering\arraybackslash}X
       >{\centering\arraybackslash}X
    }
    
    \hline
    \multicolumn{1}{l|}{} &
    \multicolumn{5}{c|}{PACS} & \multicolumn{5}{c}{OfficeHome} \\
    \hline
    \multicolumn{1}{l|}{Methods} & {Art} & {Cartoon} & {Photo} & {Sketch} & \multicolumn{1}{|c|}{Avg.} & {Artistic} & {Clipart} & {Product} & {Real} & \multicolumn{1}{|c}{Avg.} \\
    \hline
    \multicolumn{1}{l|}{ResNet-18} & {77.0} & {75.9} & {96.0} & {69.2} & \multicolumn{1}{|c|}{79.5} & {58.9} & {49.4} & {74.3} & {76.2} & \multicolumn{1}{|c}{64.7} \\
    \multicolumn{1}{l|}{MMD-AE~\cite{li2018domain}} & {75.2} & {72.7} & {96.0} & {64.2} & \multicolumn{1}{|c|}{77.0} & {56.5} & {47.3} & {72.1} & {74.8} & \multicolumn{1}{|c}{62.7} \\
    \multicolumn{1}{l|}{JiGen~\cite{carlucci2019domain}} & {79.4} & {75.3} & {96.0} & {71.6} & \multicolumn{1}{|c|}{80.5} & {53.0} & {47.4} & {71.4} & {72.7} & \multicolumn{1}{|c}{61.2} \\
    \multicolumn{1}{l|}{CrossGrad~\cite{shankar2018generalizing}} & {79.8} & {76.8} & {96.0} & {70.2} & \multicolumn{1}{|c|}{80.7} & {58.4} & {49.4} & {73.9} & {75.8} & \multicolumn{1}{|c}{64.4} \\
    
    \multicolumn{1}{l|}{MASF~\cite{dou2019domain}} & {80.2} & {77.1} & {94.9} & {71.6} & \multicolumn{1}{|c|}{81.0} & {-} & {-} & {-} & {-} & \multicolumn{1}{|c}{-} \\
    \multicolumn{1}{l|}{Epi-FCR~\cite{li2019episodic}} & {82.1} & {77.0} & {93.9} & {73.0} & \multicolumn{1}{|c|}{81.5} & {-} & {-} & {-} & {-} & \multicolumn{1}{|c}{-} \\
    \multicolumn{1}{l|}{EISNet~\cite{wang2020learning}} & {81.8} & {76.4} & {95.9} & {74.3} & \multicolumn{1}{|c|}{82.1} & {-} & {-} & {-} & {-} & \multicolumn{1}{|c}{-} \\

    \multicolumn{1}{l|}{L2A-OT~\cite{zhou2020learning}} & {83.3} & {78.2} & {\underline{96.2}} & {73.6} & \multicolumn{1}{|c|}{82.8} & {\underline{60.6}} & {50.1} & {\underline{74.8}} & {\textbf{77.0}} & \multicolumn{1}{|c}{65.6} \\
    
    \multicolumn{1}{l|}{SagNet~\cite{nam2021reducing}} & {83.5} & {77.6} & {95.4} & {76.3} & \multicolumn{1}{|c|}{83.2} & {60.2} & {45.3} & {70.4} & {73.3} & \multicolumn{1}{|c}{62.3} \\
    \multicolumn{1}{l|}{SelfReg~\cite{kim2021selfreg}} & {82.3} & {78.4} & {\underline{96.2}} & {77.4} & \multicolumn{1}{|c|}{83.6} & {-} & {-} & {-} & {-} & \multicolumn{1}{|c}{-} \\

    \multicolumn{1}{l|}{MixStyle~\cite{zhou2021domain}} & {84.1} & {78.8} & {96.1} & {75.9} & \multicolumn{1}{|c|}{83.7} & {58.7} & {53.4} & {74.2} & {75.9} & \multicolumn{1}{|c}{65.5} \\
    
    \multicolumn{1}{l|}{L2D~\cite{wang2021learning}} & {81.4} & {79.5} & {95.5} & {80.5} & \multicolumn{1}{|c|}{84.2} & {-} & {-} & {-} & {-} & \multicolumn{1}{|c}{-} \\

    \multicolumn{1}{l|}{FACT~\cite{xu2021fourier}} & {\underline{85.3}} & {78.3} & {95.1} & {79.1} & \multicolumn{1}{|c|}{84.5} & {60.3} & {54.8} & {74.4} & {\underline{76.5}} & \multicolumn{1}{|c}{\underline{66.5}} \\
    \multicolumn{1}{l|}{DSON~\cite{seo2020learning}} & {84.6} & {77.6} & {95.8} & {\underline{82.2}} & \multicolumn{1}{|c|}{85.1} & {59.3} & {45.7} & {71.8} & {74.6} & \multicolumn{1}{|c}{62.9} \\
    \multicolumn{1}{l|}{RSC~\cite{huang2020self}} & {83.4} & {\underline{80.3}} & {95.9} & {80.8} & \multicolumn{1}{|c|}{85.1} & {58.4} & {47.9} & {71.6} & {74.5} & \multicolumn{1}{|c}{63.1} \\
    \multicolumn{1}{l|}{StyleNeophile~\cite{StyleNeophile}} & {84.4} & {79.2} & {94.9} & {\textbf{83.2}} & \multicolumn{1}{|c|}{\underline{85.4}} & {59.5} & {\underline{55.0}} & {73.5} & {75.5} & \multicolumn{1}{|c}{65.8} \\
    \hline
    \multicolumn{1}{l|}{\ccol Ours} & {\ccol \textbf{85.6}} & {\ccol \textbf{84.2}} & {\ccol \textbf{96.5}} & {\ccol 79.1} & \multicolumn{1}{|c|}{\ccol \textbf{86.4}} & {\ccol \textbf{60.8}} & {\ccol \textbf{57.1}} & {\ccol \textbf{75.3}} & {\ccol \underline{76.5}} & \multicolumn{1}{|c}{\ccol \textbf{67.4}} \\

    \hline
    \end{tabularx}
    \label{tab:cr_pacs_officehome}
\end{table}

\noindent \textbf{Single-source domain generalization}
\label{single_src_dg}
Thanks to the simple design of our proposed method, which does not explicitly require domain labels, our method can be transparently incorporated with single-source domain generalization where we only have access to a single source domain during training. Therefore, to further evaluate the impact of our method on single-source domain generalization, our model is trained on each single domain of PACS and evaluated on the remaining target domains.

\noindent \textbf{Results.} As shown in Table.~\ref{tab:single_source}, our model, on average, significantly outperforms other baselines by 8.7\% in average accuracy. Besides, in all cases except for the case of $C \rightarrow S$, our model shows its superiority in performance. We believe this interesting result stems from the fact that our method is still able to help the model converge to flat minima and exploit the fine-grained relations between intra-domain samples even if only a single source domain is given.

\begin{table*}[!t]
    \centering
    \caption{
        Single-source domain generalization accuracy (\%) on PACS with a ResNet-18. (A: Art Painting, C: Cartoon, S:Sketch, P:Photo).
    }
    \fontsize{7}{9}\selectfont
    \begin{tabularx}{1.0\textwidth}{
    >{\centering\arraybackslash}X
    >{\centering\arraybackslash}X
    >{\centering\arraybackslash}X
    >{\centering\arraybackslash}X
    >{\centering\arraybackslash}X
    >{\centering\arraybackslash}X
    >{\centering\arraybackslash}X
    >{\centering\arraybackslash}X
    >{\centering\arraybackslash}X
    >{\centering\arraybackslash}X
    >{\centering\arraybackslash}X
    >{\centering\arraybackslash}X
    >{\centering\arraybackslash}X|
    >{\centering\arraybackslash}X
    }
    \hline
    \multicolumn{1}{l|}{Methods} & {A }\shortarrow[.12cm] {C}&
     {A }\shortarrow[.12cm] {S} &  {A }\shortarrow[.12cm] {P} &  {C }\shortarrow[.12cm] {A} &  {C }\shortarrow[.12cm] {S} &  {C }\shortarrow[.12cm] {P} &  {S }\shortarrow[.12cm] {A} &  {S }\shortarrow[.12cm] {C} &  {S }\shortarrow[.12cm] {P} &  {P }\shortarrow[.12cm] {A} &  {P }\shortarrow[.12cm] {C} &  {P }\shortarrow[.12cm] {S} &  {Avg.} \\
    \hline
    
    \multicolumn{1}{l|}{ResNet-18} & 62.3 & 49.0 & 95.2 & 65.7 & 60.7 & 83.6 & 28.0  & 54.5 & 35.6 & 64.1 & 23.6 & 29.1 & 54.3 \\
    \multicolumn{1}{l|}{JiGen~\cite{carlucci2019domain}} & 57.0 & 50.0 & 96.1 & 65.3 & 65.9 & 85.5 & 26.6 & 41.1 & 42.8 & 62.4 & 27.2 & 35.5 & 54.6 \\
    \multicolumn{1}{l|}{MixStyle~\cite{zhou2021domain}} & 65.5 & 49.8 & \underline{96.7} & 69.9 & 64.5 & 85.3 & 27.1 & 50.9 & 32.6 & 67.7 & \underline{38.9} & 39.1 & 57.4 \\
    \multicolumn{1}{l|}{RSC~\cite{huang2020self}} & 62.5 & 53.1 & 96.2 & 68.9 & \textbf{70.3} & 85.8 & 37.9 & 56.3 & \underline{47.4} & 66.3 & 26.4 & 32.0 & 58.6 \\
    \multicolumn{1}{l|}{SelfReg~\cite{kim2021selfreg}} & 65.2 & 55.9 & 96.6 & 72.0 & \underline{70.0} & \underline{87.5} & 37.1 & 54.0 & 46.0 & 67.7 & 28.9 & 33.7 & 59.5 \\
    \multicolumn{1}{l|}{SagNet~\cite{nam2021reducing}} & \underline{67.1} & \underline{56.8} & 95.7 & \underline{72.1} & 69.2 & 85.7 & \underline{41.1} & \underline{62.9} & 46.2 & \underline{69.8} & 35.1 & \underline{40.7} & \underline{61.9} \\

    
    \hline

    \multicolumn{1}{l|}{\ccol Ours} & \ccol \textbf{74.6} & \ccol \textbf{58.1} & \ccol \textbf{96.8} & \ccol \textbf{74.4} & \ccol 69.6 & \ccol \textbf{87.6} & \ccol \textbf{43.3} & \ccol \textbf{65.6} & \ccol \textbf{50.3} & \ccol \textbf{71.4} & \ccol \textbf{54.3} & \ccol \textbf{51.5} & \ccol \textbf{66.5} \\
    
    \hline
    \end{tabularx}
    \label{tab:single_source}
\end{table*}

\begin{table*}[!t]
    \centering
    \caption{Domain generalization accuracy (\%) on DomainBed. 
    The column ``Terra'' stands for TerraIncognita dataset. Note that we adopt leave-one-domain-out cross-validation as a model selection criteria. 
    }
    \adjustbox{max width=1.0\textwidth}
    {    
    \fontsize{8}{10}\selectfont
    \begin{tabularx}{1.\textwidth} {
    >{\centering\arraybackslash}X
    >{\centering\arraybackslash}X
    >{\centering\arraybackslash}X
    >{\centering\arraybackslash}X
    >{\centering\arraybackslash}X
    >{\centering\arraybackslash}X
    >{\centering\arraybackslash}X|
    >{\centering\arraybackslash}X
    }
    \hline
    \multicolumn{8}{c}{Model selection: leave-one-domain-out cross-validation} \\
    \hline
    \multicolumn{1}{l|}{Methods}
    & {CMNIST} & {RMNIST} & {VLCS} & {PACS} & {OfficeHome} & {Terra} & {Avg.} \\
\hline
\multicolumn{1}{l|}{ERM~\cite{vapnik1998statistical}} & 36.7 & 97.7 & 77.2 & 83.0 & 65.7 & 41.4 & 66.9 \\
\multicolumn{1}{l|}{IRM~\cite{arjovsky2019invariant}} & 40.3 & 97.0 & 76.3 & 81.5 & 64.3 & 41.2 & 66.7 \\
\multicolumn{1}{l|}{GroupDRO~\cite{sagawa2020distributionally}} & 36.8 & 97.6 & \underline{77.9} & 83.5 & 65.2 & 44.9 & 66.7 \\
\multicolumn{1}{l|}{Mixup~\cite{zhang2017mixup}} & 33.4 & \underline{97.8} & 77.7 & 83.2 & 67.0 & \textbf{48.7} & 67.9 \\
\multicolumn{1}{l|}{MLDG~\cite{li2018mldg}} & 36.7 & 97.6 & 77.2 & 82.9 & 66.1 & 46.2 & 67.7 \\
\multicolumn{1}{l|}{CORAL~\cite{sun2016coral}} & 39.7 & \underline{97.8} & \textbf{78.7} & 82.6 & \textbf{68.5} & 46.3 & \textbf{68.9} \\
\multicolumn{1}{l|}{MMD~\cite{li2018domain}} & 36.8 & \underline{97.8} & 77.3 & 83.2 & 60.2 & 46.5 & 66.9 \\
\multicolumn{1}{l|}{DANN~\cite{ganin2016dann}} & \underline{40.7} & 97.6 & 76.9 & 81.0 & 64.9 & 44.4 & 67.5 \\
\multicolumn{1}{l|}{CDANN~\cite{li2018deep}} & 39.1 & 97.5 & 77.5 & 78.8 & 64.3 & 39.9 & 66.1 \\
\multicolumn{1}{l|}{MTL~\cite{blanchard2021domain}} & 35.0 & \underline{97.8} & 76.6 & \underline{83.7} & 65.7 & 44.9 & 67.2 \\
\multicolumn{1}{l|}{SagNet~\cite{nam2021reducing}} & 36.5 & 94.0 & 77.5 & 82.3 & \underline{67.6} & \underline{47.2} & 67.5 \\
\multicolumn{1}{l|}{ARM~\cite{zhang2020adaptive}} & 36.8 & \textbf{98.1} & 76.6 & 81.7 & 64.4 & 42.6 & 66.7 \\
\multicolumn{1}{l|}{VREx~\cite{krueger2021out}} & 36.9 & 93.6 & 76.7 & 81.3 & 64.9 & 37.3 & 65.1 \\
\multicolumn{1}{l|}{RSC~\cite{huang2020self}} & 36.5 & 97.6 & 77.5 & 82.6 & 65.8 & 40.0 & 66.6 \\
\hline
\multicolumn{1}{l|}{\ccol Ours} & \ccol \textbf{46.5} & \ccol 97.7 & \ccol 74.8 & \ccol \textbf{83.8} & \ccol 65.0 & \ccol 42.5 & \ccol \underline{68.4} \\

\hline

    \end{tabularx}
    }
    \label{tab:domainbed}
\end{table*}

\noindent \textbf{DomainBed.}
We also conduct extensive experiments on the DomainBed~\cite{gulrajani2020search} which is a testbed for domain generalization to compare state-of-the-art methods across several benchmark datasets. The rationale behind the DomainBed is that the domain generalization performances are too much dependent on the hyperparameter tuning. For a fair comparison, we follow its standard protocols for training and evaluation.

\noindent \textbf{Results.} As shown in Table.~\ref{tab:domainbed}, our method generally shows competitive performances and ranks second out of 15 methods on average accuracy. In particular, on CMNIST, our method substantially outperforms other competing methods. Since CMNIST is designed to simulate the domain shift by correlating the digit colors with the class labels, we conjecture that our improvement on CMNIST is attributed to the de-stylization effect of UniStyle, which would help the model decorrelate between the colors and labels.

\subsection{Generalization in person re-ID}
\label{dg_reid}
In this section, we further evaluate our method on person re-identification (re-ID), which is to match pedestrians across non-overlapping camera views.


\noindent \textbf{Experimental setup.}
Here, we address domain generalization for person re-ID, where the test data is collected from cameras of the unseen dataset rather than from those of the training dataset. Specifically, the model trained to match people in the source dataset is evaluated by how well it matches pedestrian data of the unseen test set, which are disjoint from the source dataset. For datasets, we adopt two widely-used benchmarks: Market1501 (Market)~\cite{zheng2015scalable} and DukeMTMC-reID (Duke)~\cite{ristani2016performance,zheng2017unlabeled}. We use 32,668 images of 1,501 identities collected from 6 cameras and 36,411 images of 1,812 identities from 8 cameras for Market1501 and Duke, respectively. As for performance measures, we adopt mean average precision (mAP) and Recall@K (R@K). Following the prior work~\cite{zhou2021domain}, we adopt ResNet-50~\cite{resnet} as a backbone architecture. In these experiments, we apply UniStyle to the 1st, 2nd, and 3rd residual blocks of a model.

\begin{table}[!t]
    \centering
    \caption{
        Generalization results on the cross-dataset person re-ID.
    }
    \fontsize{8}{10}\selectfont
    \begin{tabularx}{0.75 \textwidth}{
       >{\centering\arraybackslash}X|
       >{\centering\arraybackslash}X
       >{\centering\arraybackslash}X|
       >{\centering\arraybackslash}X
       >{\centering\arraybackslash}X}
    
    \hline
    \multicolumn{1}{l|}{} &
    \multicolumn{2}{c|}{Market $\rightarrow$ Duke} & \multicolumn{2}{c}{Duke $\rightarrow$ Market} \\
    \multicolumn{1}{l|}{Methods} & {mAP} & {R@1} &  {mAP} & {R@1} \\
    
    \hline
    
    
    

    \multicolumn{1}{l|}{ResNet-50} & 19.3 & 35.4 & 20.4 & 45.2\\
    \multicolumn{1}{l|}{RandomErase~\cite{zhong2020random}} & 14.3 & 27.8 & 16.1 & 38.5\\
    \multicolumn{1}{l|}{DropBlock~\cite{ghiasi2018dropblock}} & 18.2 & 33.2  & 19.7 & 45.3\\
    
    \multicolumn{1}{l|}{MixStyle~\cite{zhou2021domain}} & 23.4 & 43.3 & 24.7 & 53.0 \\
    
    \multicolumn{1}{l|}{StyleNeophile~\cite{StyleNeophile}} & \underline{26.3} & \underline{46.5} & \underline{27.2} & \underline{55.0} \\
    
    
    
    \multicolumn{1}{l|}{\ccol Ours} & \ccol \textbf{27.4} & \ccol \textbf{49.3} & \ccol \textbf{30.1} & \ccol \textbf{59.0} \\

    \hline
    
    \end{tabularx}
    \label{tab:person_reid}
\end{table}
\noindent \textbf{Comparison to other regularization methods.}
As shown in Table.~\ref{tab:person_reid}, our method substantially outperforms other methods in mAP and Recall@1. Although RandomErase and Dropblock are effective for learning discriminative features, they fail to improve performance when encountering unseen domain data. Furthermore, by exploiting inter-class relations provided by different cameras, our method shows its superiority over MixStyle and StyleNeophile which are designed for domain generalization but utilizes one-hot labels only.

\begin{table}[!t]
    \centering
    \caption{
    mIoU (\%) results on the cross-dataset semantic segmentation. GTA5 is for training, and Cityscapes, SYNTHIA, BDD, and Mapillary are test sets.
    }
    \fontsize{8}{10}\selectfont
    \begin{tabularx}{0.9 \textwidth}{
       >{\centering\arraybackslash}X
       >{\centering\arraybackslash}X
       >{\centering\arraybackslash}X
       >{\centering\arraybackslash}X
       >{\centering\arraybackslash}X
       >{\centering\arraybackslash}X
       >{\centering\arraybackslash}X}
    \hline
    \multicolumn{1}{l|}{Methods (GTA5)} & Cityscapes & BDD & Mapillary & SYNTHIA \\
    \hline
    \multicolumn{1}{l|}{DeepLabV3+~\cite{chen2018encoder}} & 28.9 & 25.1 & 28.1 & 26.2 \\
    \multicolumn{1}{l|}{SW~\cite{pan2019switchable}} & 29.9 & 27.4 & 29.7 & 27.6 \\
    \multicolumn{1}{l|}{DRPC~\cite{yue2019domain}} & \underline{37.4} & 32.1 & 34.1 & \underline{28.0} \\
    \multicolumn{1}{l|}{RobustNet~\cite{choi2021robustnet}} & 36.5 & \textbf{35.2} & \textbf{40.3} & \textbf{28.3} \\
    \multicolumn{1}{l|}{\ccol Ours} & \ccol \textbf{39.2} & \ccol \underline{32.4} & \ccol \underline{37.1} & \ccol \underline{28.0} \\
    \hline
    \end{tabularx}
    \label{tab:dg_segmentation}
\end{table}

\subsection{Generalization in semantic segmentation}
\label{dg_segmentation}
\noindent \textbf{Experimental setup.} Lastly, to investigate whether our method can be extended to the dense prediction task, evaluation on semantic segmentation is addressed here. Following the mainstream protocol, we train models on a synthetic dataset and evaluate them on several datasets which mainly belong to real-world. Specifically, we adopt GTA5~\cite{richter2016playing} as a source dataset which consists of 24,966 images. For target datasets, Cityscapes~\cite{cityscapes}, BDD~\cite{yu2018bdd100k}, and Mapillary~\cite{neuhold2017mapillary} are real-world datasets whose image sizes are 5,000, 10,000, and 25,000, respectively. Lastly, SYNTHIA~\cite{ros2016synthia} has 9,400 images. Note that ResNet-50 is used as the backbone and the common 19 classes are used across all datasets.

\noindent \textbf{Results.}
We remark that XDED constructs an ensemble by simply averaging all the logits from the pixels whose gt is the same in a mini-batch. As shown in Table~\ref{tab:dg_segmentation}, ours outperforms the competing methods overall, even if those are dedicated to this task only. We show that our method can be extended to the pixel-wise classification with little modification on XDED. Also, the results support our claim that our method is simple yet effective in a wide range of tasks.

\subsection{In-depth Analysis}
\label{sec:experiment_ablation}
\begin{table}[!t]
\centering
\caption{
Ablation study of the proposed components on cross-domain tasks of image classification (Accuracy) and person re-ID (mAP).
}
\fontsize{8}{10}\selectfont
\begin{tabularx}{0.8 \textwidth}{
   >{\centering\arraybackslash}X
   >{\centering\arraybackslash\hsize=.5\hsize}X
   >{\centering\arraybackslash\hsize=.5\hsize}X|
   >{\centering\arraybackslash}X}
\hline

\multicolumn{1}{l|}{\multirow{1}{*}{Methods}}& \multicolumn{1}{c}{Art} & \multicolumn{1}{c|}{Clipart} & {Market $\rightarrow$ Duke}\\ 



\hline
\multicolumn{1}{l|}{Vanilla}  & 77.0 & 49.4 & 19.3 \\
\multicolumn{1}{l|}{w/ UniStyle}  & 81.2 & 50.4 & \underline{26.2} \\
\multicolumn{1}{l|}{w/ XDED}  & \underline{83.3} & \underline{55.2} & 24.2 \\
\multicolumn{1}{l|}{\ccol Ours}  & \ccol \textbf{85.6} & \ccol \textbf{57.1} & \ccol \textbf{27.4} \\

\hline

\end{tabularx}
\label{tab:ablation_components}
\end{table}
\noindent \textbf{Ablation study.} To investigate the impact of each component in our method, we conduct an ablation study which is summarized in Table~\ref{tab:ablation_components}. The result reveals that two components are complementary and consistently help the model improve the generalization ability. For image classification, XDED contributes most to the performance, and UniStyle boosts the effect of XDED. On both domains, XDED uniformly improves the vanilla method by about 6\%, whereas UniStyle shows different degrees of improvement. It is because the image style discrepancy between domains in OfficeHome is less severe than that in PACS. Interestingly, for the task of person re-ID, UniStyle reveals more impact than does XDED. Due to the inherent characteristics of the task itself, the effect of XDED on collecting meaningful knowledge of the same pedestrian from different cameras may become less significant.

\begin{figure}[t]
    \centering
    \includegraphics[width=0.92\textwidth]{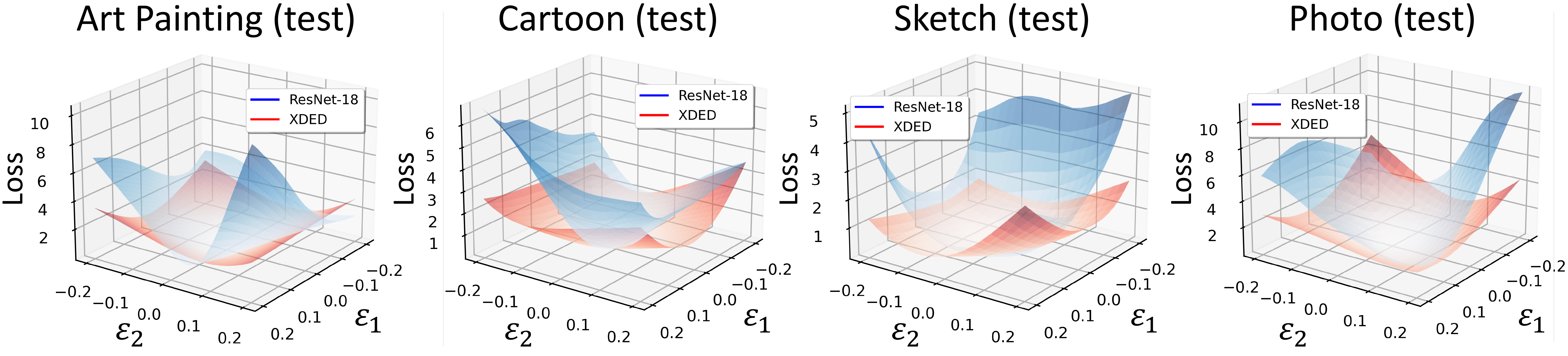}
    \caption{Visualization results of the loss landscapes incorporating the vanilla method and XDED on the PACS dataset. Note that each loss landscape is visualized on the data of source domains, not the data of the marked target domain. Blue and red surfaces are from the vanilla method and XDED, respectively.
    }
    \label{fig:loss_surface}
\end{figure}

\noindent \textbf{Loss surface visualization.}
To further illustrate how XDED leads to flat minima in the loss landscapes, we provide qualitative results that visualize the loss landscapes. Following \cite{cha2020cpr}, we plot the loss landscapes on data of source domains per each case by perturbing the model parameters across the first and second Hessian eigenvectors which are provided by PyHessian~\cite{yao2020pyhessian} which is a framework for Hessian-based analysis of neural networks. As shown in Fig.~\ref{fig:loss_surface}, we observe that the loss landscapes incorporating XDED clearly become flatter than those incorporating the vanilla method for all cases. We argue that these qualitative results also consistently support that XDED promotes flat minima.


\begin{table}[!t]
    \centering
    \caption{
    Multi-source domain generalization accuracy (\%) on Photo of PACS before and after applying given adversarial attacks.
    }
    \fontsize{8}{10}\selectfont
    \begin{tabularx}{0.75 \textwidth}{
       >{\centering\arraybackslash}X
       >{\centering\arraybackslash}X
       >{\centering\arraybackslash}X
       >{\centering\arraybackslash}X
       >{\centering\arraybackslash}X}
    \hline
    \multicolumn{1}{p{33mm}|}{Methods} & Photo & w/ FGSM & w/ PGD \\ 
    \hline
    \multicolumn{1}{l|}{ResNet-18} & 96.0 & 39.6 & 16.3\\
    \multicolumn{1}{l|}{Label smoothing~\cite{szegedy2016rethinking}} & 95.6 & 43.5 & 20.2\\
    \multicolumn{1}{l|}{Mixup~\cite{zhang2017mixup}} & 95.8 & 46.5 & 21.9\\
    \multicolumn{1}{l|}{Manifold mixup~\cite{verma2019manifold}} & 93.5 & \underline{46.6} & \underline{23.8}\\

    \multicolumn{1}{l|}{MixStyle~\cite{zhou2021domain}} & \underline{96.1} & 41.4 & 22.7\\

    \multicolumn{1}{l|}{ \ccol Ours} & \ccol \textbf{96.5} & \ccol \textbf{55.4} & \ccol \textbf{30.4}\\

    \hline
    \end{tabularx}
    \label{tab:comparison_adversarial_attack}
    \vspace{-2mm}
\end{table}
\noindent \textbf{Robustness to adversarial examples.}
Recent studies have demonstrated that convergence on flat minima strengthens the adversarial robustness~\cite{wu2020adversarial,stutz2021relating}. To revalidate that our method promotes flat minima, we evaluate the adversarial robustness of learned models. Specifically, we trained models on source domains and added adversarial perturbations on images of the unseen target domain by using existing adversarial attack methods: FGSM~\cite{goodfellow2014explaining} and PGD~\cite{madry2018towards}. Table~\ref{tab:comparison_adversarial_attack} shows that our method outperforms other regularization methods in terms of robustness against both unseen data and adversarial attacks. Considering that adversarial attacks are made to maximize the loss value, we argue that our superiority in adversarial robustness is also attributed to the capability of promoting flat minima as desired, even though our method has no direct connection to adversarial training.

\begin{table}[!t]
    \centering
    \caption{
    Average classification error (\%) on the corruption benchmarks.
    }
    \fontsize{8}{10}\selectfont
    \begin{tabularx}{0.7 \textwidth}{
       >{\centering\arraybackslash}X
       >{\centering\arraybackslash}X
       >{\centering\arraybackslash}X
       >{\centering\arraybackslash}X}
    \hline
    \multicolumn{1}{p{21mm}|}{Methods} & CIFAR-10-C & CIFAR-100-C \\
    \hline
    \multicolumn{1}{l|}{40-2 WRN~\cite{zagoruyko2016wide}} & 26.9 & 53.3 \\
    \multicolumn{1}{l|}{Cutout~\cite{devries2017improved}} & 26.8 & 53.5 \\
    \multicolumn{1}{l|}{Mixup~\cite{zhang2017mixup}} & 22.3 & 50.4 \\
    \multicolumn{1}{l|}{CutMix~\cite{yun2019cutmix}} & 27.1 & 52.9 \\
    \multicolumn{1}{l|}{AutoAug~\cite{cubuk2018autoaugment}} & 23.9 & 49.6 \\
    \multicolumn{1}{l|}{AugMix~\cite{hendrycks2019augmix}} & \textbf{11.2} & \textbf{35.9} \\

    \multicolumn{1}{l|}{ \ccol Ours} & \ccol \underline{18.5} & \ccol \underline{46.6} \\
    \hline
    \end{tabularx}
    \label{tab:comparison_corruption}
\end{table}
\noindent \textbf{Results on corruption benchmarks.} We further measure the resilience of learned models to image corruptions. Following the protocol provided by \cite{hendrycks2019benchmarking}, we trained models on the original training dataset, and evaluated them on the test dataset constructed by corrupting the original test dataset through predefined corruption types. Table.~\ref{tab:comparison_corruption} shows that our method outperforms all regularization methods except AugMix~\cite{hendrycks2020augmix}. Considering AugMix is a state of the art that is dedicated to corruption robustness while ours is not, we argue that our method still shows its significant robustness against image corruptions.
\section{Conclusion}
We have presented a simple yet effective framework for domain generalization. XDED first generates an ensemble of output distributions for the data with the same label but from different domains, and then penalizes each output distribution for the mismatch with the ensemble in the form of self-knowledge distillation. With this approach, our model can learn domain-invariant features and also easily converges to flat minima. Besides, the proposed UniStyle suppresses domain-specific style bias to boost the effect of XDED and encourage style-consistent predictions. Furthermore, we empirically validate the generalization ability of the proposed method from the perspective of flat minima and reduced divergence between source and target. Through extensive experimental results, we demonstrate the superiority of the proposed framework. 


\vfill
{\small
\noindent \textbf{Acknowledgement.} 
This work was supported by 
the NRF grant and  
the IITP grant 
funded by Ministry of Science and ICT, Korea
(NRF-2021R1A2C3012728,  
 IITP-2019-0-01906,     
 IITP-2022-0-00926,     
 IITP-2022-0-00290).    
}


%
%
\bibliographystyle{splncs04}
\bibliography{cvlab_kwak}
\end{document}


\renewcommand{\theequation}{a\arabic{equation}}
\renewcommand{\thetable}{a\arabic{table}}
\renewcommand{\thefigure}{a\arabic{figure}}
\pagestyle{headings}
\mainmatter
\def\ECCVSubNumber{4593}
\title{Cross-Domain Ensemble Distillation \linebreak for Domain Generalization}

\titlerunning{Cross-Domain Ensemble Distillation for Domain Generalization}
%
\author{Kyungmoon Lee
\inst{1,2}
Sungyeon Kim\inst{1}
Suha Kwak\inst{1}}
\authorrunning{Kyungmoon Lee, Sungyeon Kim, Suha Kwak}
%
\institute{$^{1}$POSTECH, Pohang, Korea \qquad $^{2}$NALBI Inc, Seoul, Korea \\
}

\maketitle
\appendix

This supplementary material presents additional experimental results, further analyses, and implementation details, all of which are omitted from the main paper due to the space limit. In Sec.~\ref{sec:cr_supple_exp}, additional experimental results of our method are provided. Sec.~\ref{sec:cr_supple_xded} and Sec.~\ref{sec:cr_supple_unistyle} present further analyses of the proposed components, respectively. Lastly, implementation details are in Sec~\ref{sec:cr_supple_implementation}.



\section{Additional experimental results}
\label{sec:cr_supple_exp}


\subsection{Comparison to existing self-KD methods.}
To further assess the impact of the ensemble distillation, we also compare XDED with state-of-the-art self-knowledge distillation (self-KD) methods in spite of sharing a similar methodology. Furthermore considering that our method does not explicitly depend on domain labels, we investigate the impact of XDED on the standard benchmark datasets for the image recognition task, which assume the independent and identically distributed (i.i.d.) condition. As shown in Table~\ref{tab:comparison_iid}, XDED outperforms not only the vanilla method but also other self-KD methods for all datasets. Considering that XDED is not dedicated to the case under the i.i.d. condition whereas other methods are dedicated to the case under the i.i.d. condition, these experimental results support our view that XDED can provide more meaningful supervisory signals thanks to the use of ensemble which encodes the complementary knowledge from the data with the same label.

\subsection{Results on input deformations}
\begin{figure*}[t]
    \centering
    \includegraphics[width=1.0\textwidth]{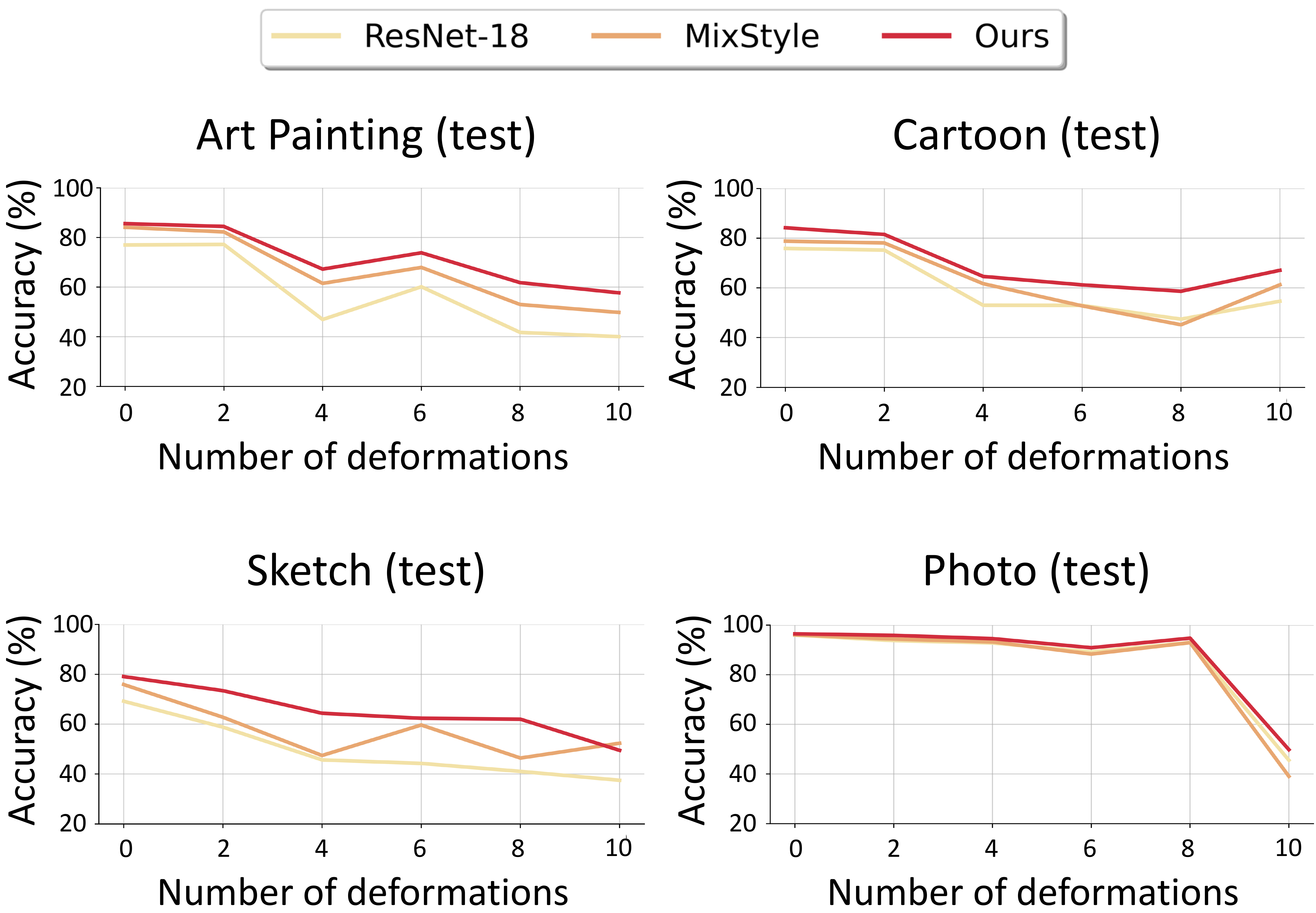}
    \caption{Test accuracy (\%) on target domains with input deformations. All models are trained under the multi-source domain generalization setting.
    }
    \label{fig:supple_deformation}
\end{figure*}

To further assess the quality of the robustness to unseen domains, we measure the changes in test accuracy as the domain gap becomes larger. We simulate larger domain gaps by increasing the degree of input deformations. Specifically, we applied multiple augmentations defined in RandAugment~\cite{cubuk2020randaugment} to images of the unseen target domain (\ie, Cartoon of PACS), and gradually increased the number of deformations\footnote{For a fair comparison, we use the same operations for each setting.}. Similar to the result of table 9 in our main paper, Fig.~\ref{fig:supple_deformation} shows that our method outperforms baseline methods against pixel-level deformations on unseen domains.

\begin{table}[!t]
    \centering
    \caption{
    Accuracy (\%) on general image recognition benchmarks.
    }
    \begin{tabularx}{0.8 \textwidth}{
       >{\centering\arraybackslash}X
       >{\centering\arraybackslash}X
       >{\centering\arraybackslash}X
       >{\centering\arraybackslash}X}
    \hline
    \multicolumn{1}{l|}{Methods} & CIFAR-10 & CIFAR-100 \\
    \hline
    
    \multicolumn{1}{l|}{ResNet-18} & 94.6 & 75.2 \\
    
    \multicolumn{1}{l|}{CS-KD~\cite{yun2020regularizing}} & 94.6 & 78.0 \\
    \multicolumn{1}{l|}{BYOT~\cite{zhang2019your}} & \underline{95.1} & \underline{78.6} \\
    
    \multicolumn{1}{l|}{ \ccol XDED} & \ccol \textbf{95.3} & \ccol \textbf{78.7} \\
    \hline
    \end{tabularx}
    
    \label{tab:comparison_iid}
\end{table}

\subsection{Learning acceleration}
\label{sec:supple_convergence}
\begin{figure}[!t]
    \centering
    \includegraphics[width=0.8\columnwidth]{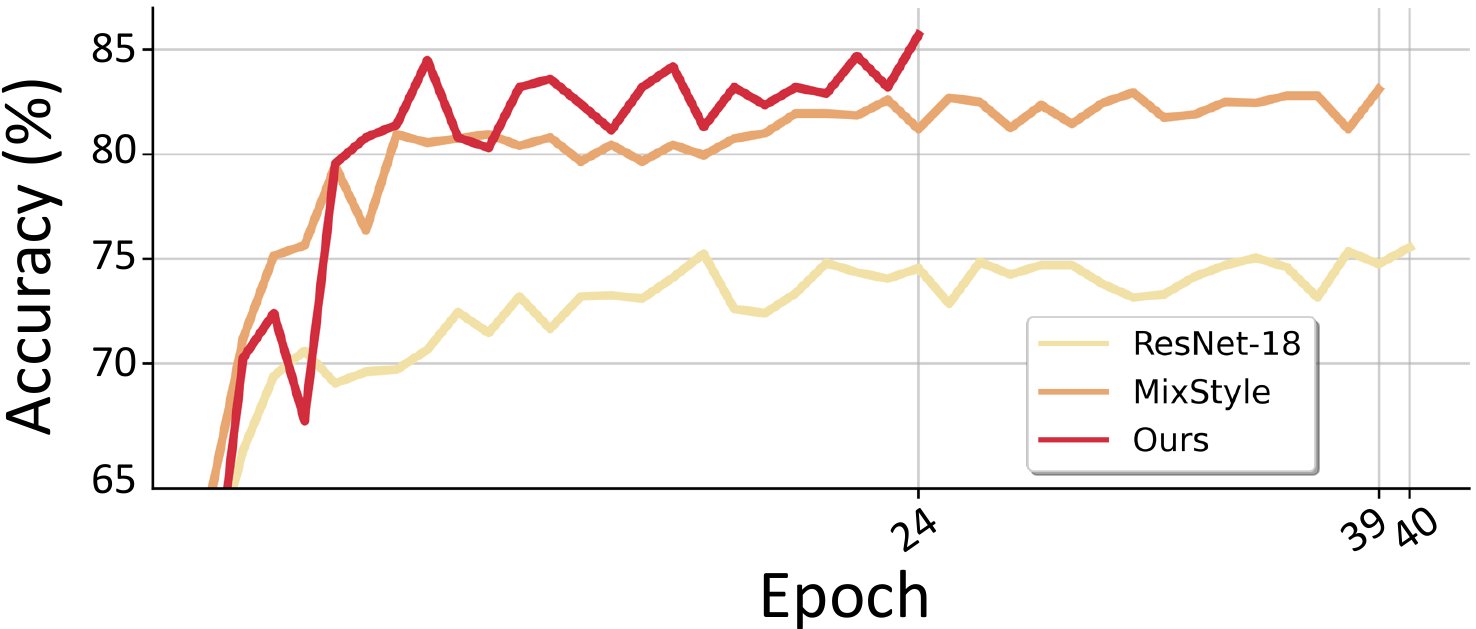}
    \caption{Multi-source domain generalization accuracy (\%) versus the number of epochs on the unseen target domain, Art Painting of PACS. Note that all models were trained on the rest three domains of PACS (\ie, Cartoon, Sketch, and Photo) and all methods were trained with the same hyperparameter setting (\eg, batch size and learning rate) on a single RTX 2080 GPU.
    }
    \label{fig:supple_convergence}
\end{figure}

Our method also enables faster convergence of the learning process. In generalization on PACS, our effectiveness of learning acceleration is demonstrated. As shown in Fig.~\ref{fig:supple_convergence}, our method reaches highest performance with less epochs (\ie, 24 epochs) compared to both vanilla method (40 epochs) and MixStyle~\cite{zhou2021domain} (39 epochs). Although the vanilla method requires many iterations to address the domain gaps between source domains, our method accelerates the learning process via promoting flat minima which produce lower errors in not only source domains but also the target domain. Since MixStyle~\cite{zhou2021domain}, as an augmentation method, synthesizes novel styles via mixing statistics at the feature level, it is inherently limited to requiring as many epochs as possible to address many augmented styles for its best performance. Also, in the context of the actual training time in total, compared to MixStyle, we remark that our method significantly reduces the training complexity. In detail, our method increases training time per iteration by 5\% but requires only 60\% of training epochs and achieves about 1.6 times more performance boost averaged over every target domain of the PACS dataset. The results are summarized in Table~\ref{tab:supple_convergence}.

\begin{table}[!t]
    \centering
    \caption{ Performance boosts over training complexity.
    }
    \begin{tabularx}{0.7\textwidth}{ 
           >{\centering\arraybackslash}X |
           >{\centering\arraybackslash}X |
           >{\centering\arraybackslash}X} 
    \hline
    \multicolumn{1}{l|}{Methods} & Millisec / iter & Accuracy boost (\%) \\
    \hline
    
    
    \multicolumn{1}{l|}{ResNet-18} & 34 & -\\
    \multicolumn{1}{l|}{MixStyle~\cite{zhou2021domain}} & 35 & 4.2 \\
    \multicolumn{1}{l|}{\ccol Ours} & \ccol 37 & \ccol 6.9\\
    
    \hline
    \end{tabularx}
    \label{tab:supple_convergence}
\end{table}

\begin{figure}[t]
    \centering
    \includegraphics[width=0.9\textwidth]{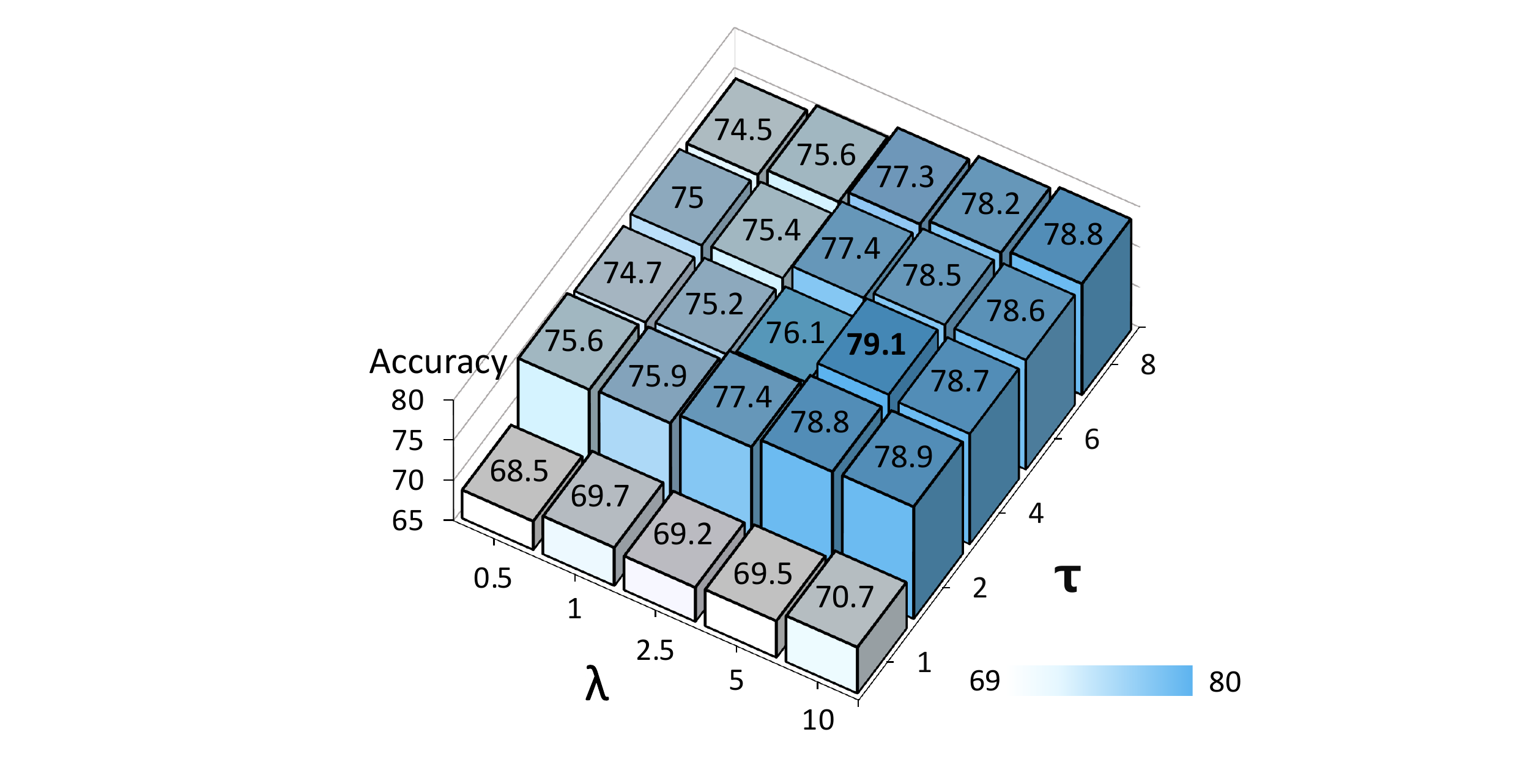}
    \caption{Multi-source domain generalization accuracy (\%) versus hyperparameters $\lambda$ and $\tau$ on Sketch of the PACS dataset.
    }
    \label{fig:supple_hyperparam}
\end{figure}

\subsection{Impact of hyperparameters}
We empirically investigated the effect of the two hyperparameters of XDED: $\lambda$ and $\tau$. We examine accuracy of multi-source domain generalization on Sketch of the PACS dataset by varying the values of $\lambda \in \{ 0.5, 1.0, 2.5, 5.0, 10.0 \} $ and $\tau \in \{ 1.0, 2.0, 4.0, 6.0, 8.0 \} $. The results are summarized in Fig.~\ref{fig:supple_hyperparam}. First, the performance boost is the highest when $ \tau $ is increased from 1.0 to 2.0. However, if $\tau$ exceeds 2.0, the performance boost is somewhat reduced. Next, $\lambda$ also affects the performance. The performance boost commonly increases with the value of $\lambda$. However, the degree of boost is also reduced when $\tau$ is 1.0. To sum it up, the accuracy of our method is generally high and stable when $\lambda$ and $\tau$ are greater than 5.0 and 2.0, and we argue that this result supports our view that our method does not strongly depend on the specific hyperparameter setting.

\section{Additional analysis results on XDED}
\label{sec:cr_supple_xded}
\begin{figure*}
    \centering
    \includegraphics[width=1.0\textwidth]{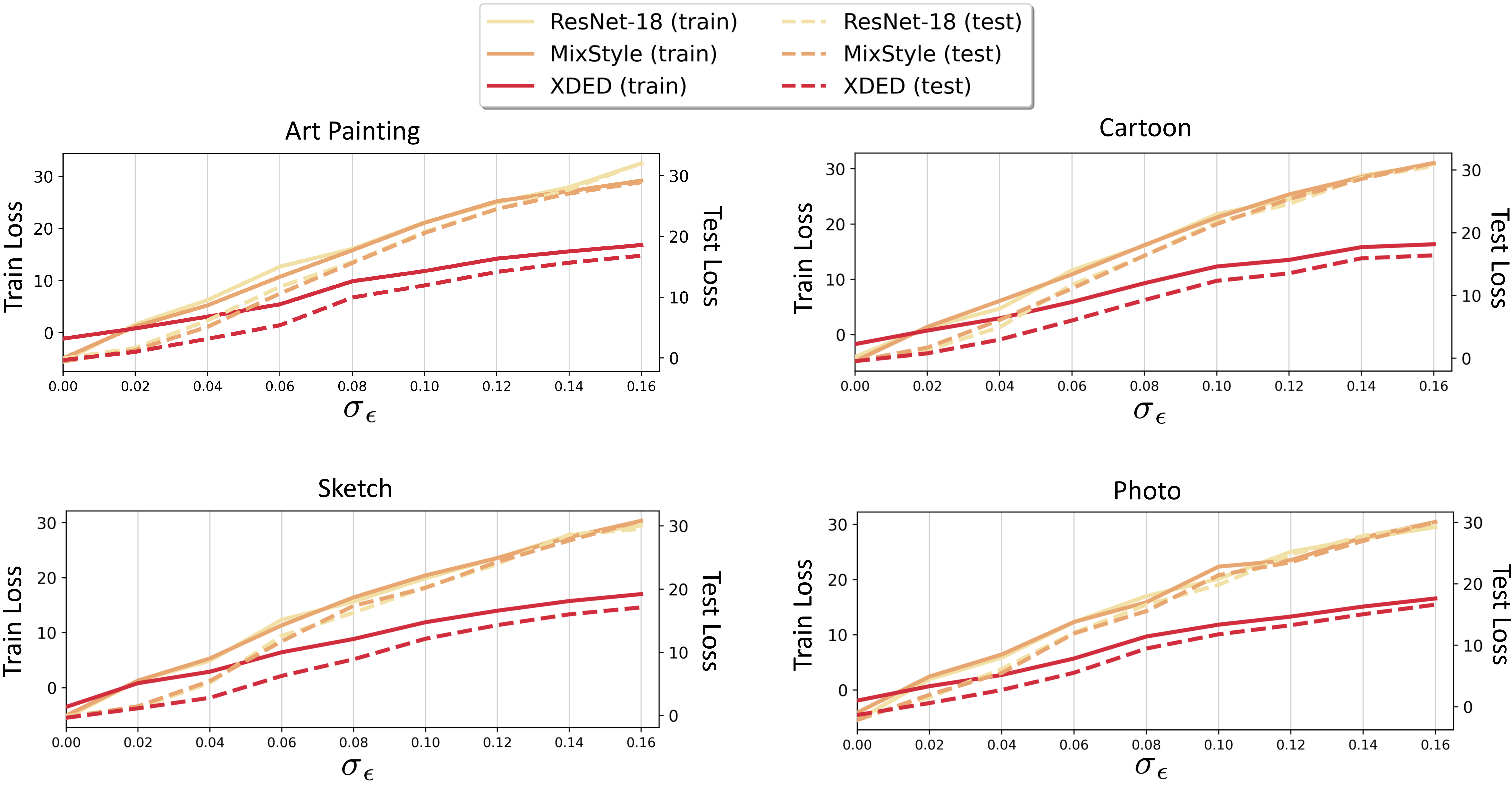}
    \caption{Train/Test losses versus the weight perturbation while varying the standard deviation of the added Gaussian noise. Note that train losses are calculated over the remaining three source domains for each target domain and the loss values are log-scaled.
    }
    \label{fig:supple_weight_perturbation}
\end{figure*}

In this section, we provide further analyses of XDED. Also, we provide additional empirical results on the flatness of local minima (Fig.~\ref{fig:supple_weight_perturbation}).

\begin{table}[!t]
    \centering
    \caption{
    Generalization results on the target domain (Clipart of OfficeHome) and required overhead during training.
    }
    \begin{tabularx}{0.7 \textwidth}{
       >{\centering\arraybackslash}X
       >{\centering\arraybackslash}X
       >{\centering\arraybackslash}X
       >{\centering\arraybackslash}X
       >{\centering\arraybackslash}X}
    \hline
    \multicolumn{1}{l|}{Methods} & Vanilla & SAM~\cite{foret2021sharpness}  & \ccol XDED  \\
    \hline
    \multicolumn{1}{l|}{Accuracy (\%) ($\uparrow$)} & 49.4 & \underline{51.9} & \ccol \textbf{55.2} \\
    \multicolumn{1}{l|}{Time / iter (ms) ($\downarrow$)} & \textbf{32} & 72 & \ccol \underline{33} \\
    \hline
    \end{tabularx}
    \label{tab:supple_flat_minima_cost}
\end{table}
\subsection{Convergence into flat minima at a low cost}
Since there are several ways for convergence into flat minima, we believe that the required overhead to achieve that goal should be considered as another criterion of valuation for a practical usage. Therefore, according to this point of view, we compare XDED with SAM~\cite{foret2021sharpness} which has been proposed to optimize loss sharpness more directly. As shown in Table.~\ref{tab:supple_flat_minima_cost}, XDED shows superior robustness on the unseen domain over SAM. Furthermore, XDED requires negligible overhead while SAM requires the training time more than 2 times longer than the vanilla method and XDED. This is because SAM must additionally compute gradient approximation for the loss sharpness to be optimized, which exposes a trade-off between training complexity and the optimization of the loss sharpness. 
On the other hand, SAM's performance is inferior to that of XDED, since it directly promotes a flat minima but cannot learn domain-invariant features, unlike our method. To sum it up, we argue that XDED offers more appealing option for domain generalization in the context of both performance improvement and required overhead.

\subsection{Ensemble matters for generalization ability}
\label{sec:supple_standard_benchmarks}
\begin{figure}[!t]
    \centering
    \includegraphics[width=0.82\textwidth]{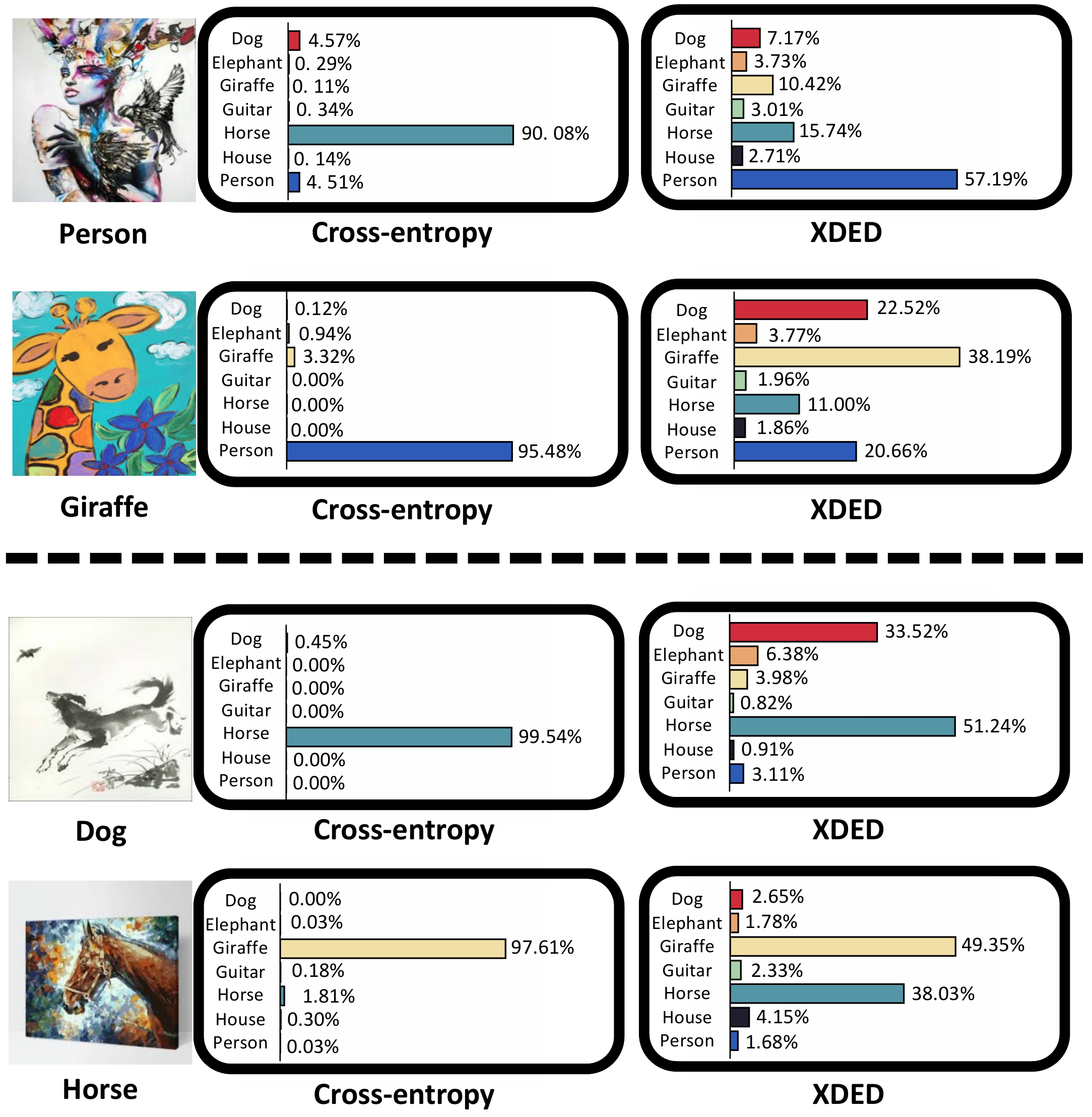}
    \caption{Prediction scores for misclassified samples from the vanilla method (\ie, Cross-entropy). \textbf{Top}: Well-classified results from XDED, \textbf{Bottom}: Misclassified results from XDED.
    }
    \label{fig:supple_qualitative}
\end{figure}
To examine how XDED affects the predictions of the learned model in detail, we investigate the softmax scores from the model which is trained on the source domains (\ie, Cartoon, Sketch and Photo of PACS) and evaluated on the target domain (\ie, Art Painting). Specifically, for the misclassified samples from the vanilla method, we assess the differences in the softmax scores from the vanilla method and XDED. First, as shown in Fig.~\ref{fig:supple_qualitative} (Top), XDED can lead to accurate predictions on the samples which the vanilla method confidently misclassifies. For the giraffe sample, the model learned with XDED not only predicts the ground truth successfully but also puts relatively high probability on the similar class (\ie, Dog) while lowering the probabilities of the irrelevant classes (\ie, Guitar and House). Similarly, even in its failure cases, XDED encourages the model to predict with consideration of inter-class relations (\ie, increasing scores for similar classes while suppressing those for irrelevant classes) as shown in Fig.~\ref{fig:supple_qualitative} (Bottom). By exploiting the ensemble incorporating information of multiple domains, XDED encourages the model to consider richer inter-class relations which may be exploited on the target domain and lead to meaningful predictions even in failure cases. Note that UniStyle is excluded from the results discussed above.

\subsection{Comparison with other ensemble methods.}
\label{sec:supple_comparison_ensemble}
Lastly, we compare XDED with other domain generalization methods which adopt ensemble techniques. The comparison analysis is summarized in Table~\ref{tab:comparison_ensemble}. We remark that the proposed method does not require any additional module and is also free from domain labels (Note that our method, thus, seamlessly extends to the single-source domain generalization setting, which is discussed in Sec.4.1 of our main paper.). Nevertheless, our method clearly outperforms other methods in the context of generalization ability. In contrast, DAEL~\cite{zhou2021dael} and DSON~\cite{seo2020learning} show inferior results even though they explicitly exploit domain labels and multiple domain-specific modules (\ie, classifiers and BN layers~\cite{Batchnorm}, respectively.). DAEL is the most similar to our method but clearly different in the sense that DAEL exploits the predictions of an expert as supervisory signals for non-experts on the same instance, but our method constructs the ensemble using multiple instances and exploits it as supervision to model itself. In other words, by learning domain-invariant information through the ensemble of meaningful knowledge contained in different domains, better performance could be achieved even with limited resources.
\begin{table}[!t]
    \centering
    \fontsize{9}{11}\selectfont
    \caption{
    Comparison with other ensemble methods which are dedicated to domain generalization. Extra $\theta$ and Domain ID denote whether each method requires additional modules and domain labels are explicitly bound to be given, respectively. Accuracy denotes the average accuracy of each method over PACS and Office-Home.
    }
    \begin{tabularx}{0.8\textwidth}{ 
           >{\centering\arraybackslash}X |
           >{\centering\arraybackslash}X |
           >{\centering\arraybackslash}X |
           >{\centering\arraybackslash}X} 
    \hline
    \multicolumn{1}{l|}{Methods} & Extra $\theta$ & Domain ID & Accuracy \\
    \hline
    \multicolumn{1}{l|}{DAEL~\cite{zhou2021dael}} & \ding{51} & \ding{51} & 74.7\\
    \multicolumn{1}{l|}{DSON~\cite{seo2020learning}} & \ding{51} & \ding{51} & 74.0\\
    \multicolumn{1}{l|}{\ccol Ours} & \ccol \ding{55} & \ccol \ding{55} & \ccol 76.9\\

    \hline
    \end{tabularx}
    \label{tab:comparison_ensemble}
\end{table}

\section{Additional analysis results on UniStyle}
\label{sec:cr_supple_unistyle}

\begin{table*}[!t]
    \caption{Ablation study on where to apply UniStyle in the ResNet architecture.}
    \footnotesize

    \begin{subtable}{.5\linewidth}
      \centering
        \caption{Image classification on PACS}
        \begin{tabular}{l c}
            \hline
            Model & Accuracy (\%) \\
            \hline
            ResNet-18 & 79.5 \\
            \hline
            + MixStyle (\textsc{res1}) & 80.1 \\
            + UniStyle (\textsc{res1}) & 81.5 \\
            \hline
            + MixStyle (\textsc{res12}) & 81.6 \\
            + UniStyle (\textsc{res12}) & \textbf{82.9} \\
            \hline
            + MixStyle (\textsc{res123}) & 82.8 \\
            + UniStyle (\textsc{res123}) & 82.4 \\
            \hline
            + MixStyle (\textsc{res1234}) & 75.6 \\
            + UniStyle (\textsc{res1234}) & 12.8 \\
            \hline
        \end{tabular}
    \end{subtable}%
    \begin{subtable}{.5\linewidth}
      \centering
        \caption{Person re-ID from Market1501$\rightarrow$Duke}
        \begin{tabular}{ll}
            \hline
            Model & mAP (\%) \\
            \hline
            ResNet-50 & 19.3 \\
            \hline
            + MixStyle (\textsc{res1}) & 22.6 \\
            + UniStyle (\textsc{res1}) & 23.3 \\
            \hline
            + MixStyle (\textsc{res12}) & 23.8 \\
            + UniStyle (\textsc{res12}) & 24.1 \\
            \hline
            + MixStyle (\textsc{res123}) & 22.0 \\
            + UniStyle (\textsc{res123}) & \textbf{26.2} \\
            \hline
            + MixStyle (\textsc{res1234}) & 10.2 \\
            + UniStyle (\textsc{res1234}) & 0.2 \\
            \hline
        \end{tabular}
    \end{subtable}
    \label{tab:ablation_where_to_whiten}
\end{table*}
\subsection{Where to apply UniStyle}
We remark that UniStyle can be applied after arbitrary intermediate layers of the backbone (\ie, ResNet~\cite{resnet}) as a plug-and-play module. Therefore, to investigate the impact of where UniStyle is applied, we evaluate the generalization performance in image classification and person re-ID while varying the locations of where the operation is applied. For a baseline, MixStyle~\cite{zhou2021domain} is adopted and compared with our UniStyle since they share the commonality of being applied to multiple intermediate feature maps.
For brevity, let \textsc{res\#} denote the indexes of residual blocks where the specified operation is applied (\eg, \textsc{res12} means the operation is applied after both the first and second residual blocks). As shown in Table.~\ref{tab:ablation_where_to_whiten}, we observe that UniStyle and MixStyle have a similar trend in both tasks. First, since early layers are known to capture low-level features such as texture or edge information, it is pertinent for both UniStyle and MixStyle to be applied after early layers to remove style bias and synthesize novel styles, respectively. However, UniStyle achieves the best performances on both tasks, which indicates the importance of removing and unifying style bias rather than augmenting novel styles. Next, on the contrary, both operations lead critical performance drop when incorporated with the last layer, \textsc{res4}, since late layers are known to address semantic information (\ie, their statistics would be highly correlated with target labels). In detail, MixStyle perturbs the statistics by interpolating those of two different instances that may have different labels, whereas UniStyle normalizes the feature map of all samples regardless of their labels, resulting in a more critical performance drop. Note that XDED is excluded in results of Table.~\ref{tab:ablation_where_to_whiten}.

\subsection{Comparison with other methods}
Even though UniStyle is not the first that addresses feature-level statistics (\ie, style information), we take notice of its appealing trade-off between performance boost and training complexity and compare it with other related methods. The results are summarized in Table~\ref{tab:supple_comparison_unistyle}. First, UniStyle outperforms SagNet~\cite{nam2021reducing} and MixStyle~\cite{zhou2021domain}, all of which are dedicated to domain generalization. This indicates that directly removing and unifying the style bias is more effective for generalization ability than augmenting feature statistics or learning style-agnostic features.
Next, although BIN~\cite{nam2018batch} and SRM~\cite{lee2019srm} adaptively recalibrate feature statistics and achieve state-of-the-art performance in image recognition tasks, they show inferior results on unseen domains. Last but not least, Deep Whitening Transformation (DWT)~\cite{cho2019image} which is the most similar to UniStyle shows the second-best performance. DWT has been proposed to encourage the feature extractor to naturally encode the whitened feature by enforcing consistency regularization between the covariance matrix of the feature and identity matrix. Nevertheless, as a constraint that is too strong, DWT weakens feature discrimination and excessively throws away content information, resulting in performance that lags behind UniStyle.
\begin{table}[!t]
    \centering
    \caption{
    Comparison between UniStyle and other related methods. Extra $\theta$ and Extra $\mathcal{L}$ denote whether each method requires additional modules and loss function, respectively, which both increase training complexity and resources. Accuracy denotes the multi-source domain generalization accuracy (\%) of each method on the target domain, Cartoon of the PACS dataset.
    }
    \begin{tabularx}{0.85\textwidth}{ 
           >{\centering\arraybackslash}X |
           >{\centering\arraybackslash}X |
           >{\centering\arraybackslash}X |
           >{\centering\arraybackslash}X} 
    \hline
    \multicolumn{1}{l|}{Methods} & Extra $\theta$ & Extra $\mathcal{L}$ & Accuracy \\
    \hline
    
    
    \multicolumn{1}{l|}{ResNet-18} & \ding{55} & \ding{55} & 75.9\\

    \hline
    \multicolumn{4}{c}{Domain Generalization}\\
    \hline

    \multicolumn{1}{l|}{SagNet~\cite{nam2021reducing}} & \ding{51} & \ding{51} & 77.6\\
    
    \multicolumn{1}{l|}{MixStyle~\cite{zhou2021domain}} & \ding{55} & \ding{55} & 78.8\\
    
    \hline
    \multicolumn{4}{c}{Style Transfer}\\
    \hline
    
    \multicolumn{1}{l|}{BIN~\cite{nam2018batch}} & \ding{51} & \ding{55} & 78.7\\
    \multicolumn{1}{l|}{SRM~\cite{lee2019srm}} & \ding{51} & \ding{55} & 78.9\\
    
    \multicolumn{1}{l|}{DWT~\cite{cho2019image}} & \ding{55} & \ding{51} & \underline{79.0}\\

    \hline
    \multicolumn{1}{l|}{\ccol UniStyle} & \ccol \ding{55} & \ccol \ding{55} & \ccol \textbf{79.2}\\
    
    \hline
    \end{tabularx}
    \label{tab:supple_comparison_unistyle}
\end{table}

\section{Implementation details}
\label{sec:cr_supple_implementation}
For the task of domain generalization in image classification, we train the models using the sgd optimizer with the cosine learning decay~\cite{loshchilov2016sgdr} and initial learning rate of $10^{-3}$. They are learned for 100 epochs. For batch construction, we use the batch size of 64, and sample 16 instances per class for the proposed XDED. For the task of person re-ID, we also train the models using the sgd optimizer with initial learning rate of 0.05.
\clearpage
{\small
\bibliographystyle{splncs04}
\bibliography{cvlab_kwak}
}